\documentclass[10pt,twocolumn,letterpaper]{article}

\usepackage{cvpr}
\usepackage{times}
\usepackage{epsfig}
\usepackage{graphicx}
\usepackage{amsmath}
\usepackage{amssymb}
\usepackage{subcaption}
\usepackage{pifont}
\usepackage{multirow}
\newcommand*{\affmark}[1][*]{\textsuperscript{#1}}

\usepackage[breaklinks=true,bookmarks=false]{hyperref}
\cvprfinalcopy 

\def\cvprPaperID{1270} 

\begin{document}

\title{Spatial Pyramid Based Graph Reasoning for Semantic Segmentation}

\author{Xia Li\affmark[1,2,4,*] \quad Yibo Yang\affmark[3,4,*] \quad Qijie Zhao\affmark[5] \quad Tiancheng Shen\affmark[3,4] \quad Zhouchen Lin\affmark[4] \quad Hong Liu\affmark[2] \\
\small \affmark[1] Zhejiang Lab \\ 
\small \affmark[2] Key Laboratory of Machine Perception, Shenzhen Graduate School, Peking University  \\
\small \affmark[3] Academy for Advanced Interdisciplinary Studies, Peking University \\
\small \affmark[4] Key Laboratory of Machine Perception (MOE), School of EECS, Peking University \\
\small \affmark[5] Wangxuan Institute of Computer Technology, Peking University \\
{\tt\small \{{ethanlee,ibo,zhaoqijie,tianchengshen,zlin,hongliu\}@pku.edu.cn}}
}

\maketitle
\let\thefootnote\relax\footnote{*: Equal contribution.}

\begin{abstract}
   The convolution operation suffers from a limited receptive filed, while global modeling is fundamental to dense prediction tasks, such as semantic segmentation. In this paper, we apply graph convolution into the semantic segmentation task and propose an improved Laplacian. The graph reasoning is directly performed in the original feature space organized as a spatial pyramid. Different from existing methods, our Laplacian is data-dependent and we introduce an attention diagonal matrix to learn a better distance metric. It gets rid of projecting and re-projecting processes, which makes our proposed method a light-weight module that can be easily plugged into current computer vision architectures. More importantly, performing graph reasoning directly in the feature space retains spatial relationships and makes spatial pyramid possible to explore multiple long-range contextual patterns from different scales. Experiments on Cityscapes, COCO Stuff, PASCAL Context and PASCAL VOC demonstrate the effectiveness of our proposed methods on semantic segmentation. We achieve comparable performance with advantages in computational and memory overhead.
\end{abstract}

\section{Introduction}

Convolutional Neural Networks (CNNs) based architectures have revolutionized a wide range of computer vision tasks \cite{he2016deep,Yang_2018_CVPR,chen2018deeplab,long2015fully}. Despite the huge success, convolutional operations suffer from a limited receptive filed, so they can only capture local information. Only with layers stacked as a deep model, can convolution networks have the ability to aggregate rich information of global context. However, it is an inefficient way since stacking local cues cannot always precisely handle long-range context relationships. Especially for pixel-level classification problems, such as semantic segmentation, performing long-range interactions is an important factor for reasoning in complex scenarios \cite{chen2018deeplab,chen2017rethinking}. For examples, it is prone to assign visually similar pixels in a local region into the same category. Meanwhile, pixels of the same object but distributed with a distance are difficult to construct dependencies.

Several approaches have been proposed to address the problem. Convolutional operations are reformulated with dilation \cite{yu2015multi} or learnable offsets \cite{dai2017deformable} to augment the spatial sampling locations. Non-local network \cite{wang2018non} and double attention network \cite{chen20182} try to introduce new interaction modules that sense the whole spatial-temporal space. They enlarge the receptive region and enable capturing long-range dependencies within deep neural networks. Recurrent neural networks (RNNs) can also be employed to perform long-range reasoning \cite{fan2018scene,shuai2018scene}. However, these methods learn global relationships implicitly and rely on dense computation. Because graph-based propagation has the potential benefits of reasoning with explicit semantic meaning stored in graph structure, graph convolution \cite{kipf2016semi} recently has been introduced into high-level computer vision tasks \cite{li2018beyond,liang2018symbolic,glore}. These methods first transform the grid-based CNN features into graph representation by projection, and then perform graph reasoning with graph convolution proposed in \cite{kipf2016semi}. Finally, these node features are re-projected back into the original space. The projection and re-projection processes try to build connections between coordinate space and interaction space, but introduce much computation overhead and damage the spatial relationships. 

\begin{figure*}[t!]
	\begin{center}
		\includegraphics[width=0.8\linewidth]{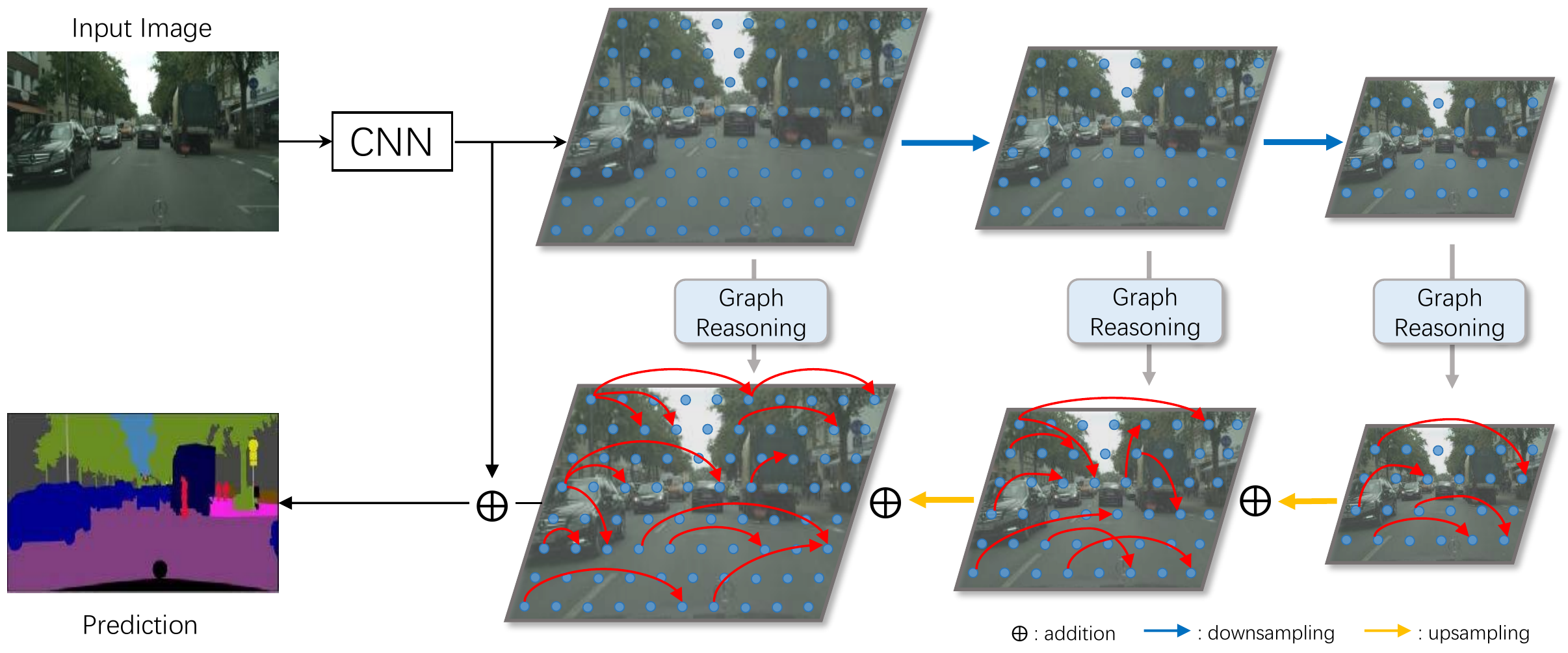}
	\end{center}
	\vspace{-5mm}
	\caption{A diagram of our model with graph reasoning on spatial pyramid for the semantic segmentation task. The graph reasoning is directly performed in the original feature space. Multiple long-range contextual patterns are captured from different scales.}
	\label{fig1}
	\vspace{-4mm}
\end{figure*}

As illustrated in Figure \ref{fig1}, in this paper, we propose an improved Laplacian formulation for graph reasoning that is directly performed in the original CNN feature space organized as a spatial pyramid. It gets rid of projection and re-projection processes, making our proposed method a light-weight module jointly optimized with the network training. Performing graph reasoning directly in the original feature space retains the spatial relationships and makes spatial pyramid possible to sufficiently exploit long-range semantic context from different scales. We name our proposed method as 
\textbf{S}patial \textbf{Py}ramid Based \textbf{G}raph \textbf{R}easoning (SpyGR) layer.

Initially, graph convolution was introduced to extract representation in non-Euclidean space, which cannot be handled well by current CNN architectures \cite{bruna2013spectral}. It seems that graph propagation should be performed on graph-structured data, which motivates the construction of semantic interaction space in \cite{li2018beyond,liang2018symbolic,glore}. Actually we note that image features can be regarded as a special case of data defined on a simple low-dimensional graph \cite{henaff2015deep}. When the graph structure of input is known, \emph{i.e.}, the Laplacian matrix $L$ is given, the graph convolution \cite{kipf2016semi} essentially performs a special form of Laplacian smoothing on the input, making each new vertice feature as the average of itself and connected neighbors \cite{li2018deeper}. But for the case that graph structure is not given, as seen in CNN features, the graph structure can be estimated with the similarity matrix from the data \cite{henaff2015deep}, which achieves a similar goal with the projection process adopted in \cite{li2018beyond,liang2018symbolic,glore}. Different from their work where the Laplacian is a learnable data-independent matrix, in this study, we modify the Laplacian as a data-dependent similarity matrix, and introduce a diagonal matrix that performs channel-wise attention on the inner product distance. The Laplacian ensures that the long-range context pattern to learn is dependent on the input features and not restricted as a specific one. Our method spares the computation to construct an interaction space by projecting. More importantly, it retains the spatial relationships to facilitate exploiting long-range context from multi-scale features.

Spatial pyramid contains multi-scale contextual information that is important for dense prediction tasks \cite{yu2015multi,zhao2017pyramid,lin2017feature}. For graph-structured data, multi-scale scheme is also the key to build hierarchical representation and enable the model to be invariant with scale changes \cite{ying2018hierarchical,liao2019lanczosnet}. Global context owns multiple long-range contextual patterns that can be better captured from features of different sizes. The finer representation has more detailed long-range context, while the coarser representation could provide more global relationships. Because our method is able to perform graph reasoning directly in the original feature space, it is possible to build a spatial pyramid to further extend the long-range contextual patterns that our method can model.

The SpyGR layer is light-weight and can be plugged into CNN architectures easily. It efficiently extracts long-range context without introducing much computational overhead. The contributions in this study are listed as follows:

\vspace{-2mm}
\begin{itemize}
	\item We propose an improved Laplacian formulation that is data-dependent, and introduce a diagonal matrix with position-agnostic attention on the inner product to enable a better distance metric. 
	\vspace{-2mm} 
	\item The Laplacian is able to perform graph reasoning in the original feature space, and makes spatial pyramid possible to capture multiple long-range contextual patterns. We develop a computing scheme that effectively reduces the computational overhead.
	\vspace{-2mm}
	\item Experiments on multiple datasets, including PASCAL Context, PASCAL VOC, Cityscapes and COCO Stuff, show the effectiveness of our proposed methods for the semantic segmentation task. We achieve top performance with advantages in computational and memory overhead. 
\end{itemize}

\section{Related Work}

\noindent\textbf{Semantic segmentation.}
Fully convolutional network (FCN) \cite{long2015fully} has been the basis of semantic segmentation with CNNs. Because details are important for dense classification problems, different methods are proposed to generate desired spatial resolution and keep object details. In \cite{noh2015learning}, deconvolution \cite{zeiler2011adaptive} is employed to learn finer representation from low-resolution feature maps, while SegNet \cite{badrinarayanan2017segnet} achieves this purpose using an encoder-decoder structure. U-Net \cite{ronneberger2015u} adds a skip connection between the down-sampling and up-sampling paths. RefineNet \cite{lin2017refinenet} introduce a multi-path refinement network that further exploits the finer information along the down-sampling path. 

Another stream aims to enhance multi-scale contextual information aggregation. In \cite{farabet2013learning}, input images are constructed as a Laplacian pyramid and each scale is fed into a deep CNN model. ParseNet \cite{liu2015parsenet} introduces image-level features to augment global context. DeepLabv2 \cite{chen2018deeplab} proposes the atrous spatial pyramid pooling (ASPP) module that consists of parallel dilated convolutions with variant dilation rates. PSPNet \cite{zhao2017pyramid} performs spatial pyramid pooling to collect contextual information of different scales. DeepLabv3 \cite{chen2017rethinking} employs ASPP module on image-level features to better aggregate global context. 

Other methods that model global context include formulating advanced convolution operations \cite{dai2017deformable,wang2018non,chen20182}, relying on attention mechanisms \cite{chen2016attention,zhang2018context,zhao2018psanet,fu2018dual}, and introducing Conditional Random Field (CRF) \cite{chen2014semantic,zheng2015conditional,liu2015semantic} or RNN variants \cite{liang2016semantic,fan2018scene,shuai2018scene} to build long-range dependencies. Still, it needs further efforts to explore how to model global context more efficiently, and perform reasoning explicitly with the semantic meanings.

\noindent\textbf{Graph convolution.} Graph convolution was initially introduced as a graph analogue of the convolutional operation \cite{bruna2013spectral}. Later studies \cite{defferrard2016convolutional,kipf2016semi} make approximations on the graph convolution formulation to reduce the computational cost and training parameters. It provides the basis of feature embedding on graph-structured data for semi-supervised learning \cite{kipf2016semi,li2018deeper}, node or graph classification \cite{velivckovic2017graph,ying2018hierarchical,zhang2018end}, and molecule prediction \cite{li2018adaptive}. Due to the ability of capturing global information in graph propagation, the graph reasoning is introduced for visual recognition tasks \cite{li2018beyond,liang2018symbolic,glore}. These methods transform the grid-based feature maps into region-based node features via projection. Different from these studies, our method notes that the graph reasoning can be directly performed in original feature space, once the learnable Laplacian matrix is data dependent. It spares the computation of projection and re-projection, and retains the spatial relationships in the graph reasoning process. 

\noindent\textbf{Feature pyramid.} Feature pyramid is an effective scheme to capture multi-scale context. It is widely adopted in dense prediction tasks such as semantic segmentation 
\cite{chen2018deeplab,zhao2017pyramid} and object detection \cite{lin2017feature,he2017mask}. Hierarchical representation is also shown to be useful for embedding on graph-structured data \cite{ying2018hierarchical}. Different from the pyramid pooling module in \cite{chen2018deeplab}, we build our spatial pyramid simply by down-sampling and up-sampling processes on the final predicting feature maps. We directly perform graph reasoning on each of the scale and aggregate them in order to capture sufficient long-range contextual relationships in the final prediction.

\section{Our Methods}

In this section, we first briefly introduce the background of graph convolution, and then develop our method in detail. Finally, we analyze the complexity of our method.

\subsection{Graph Reasoning on Graph Structures}

Graph convolution was introduced as an analogue of convolutional operation on graph-structured data. Given graph $\mathcal{G}=(V,E)$ and its adjacency matrix $A$ and degree matrix $D$, the normalized graph Laplacian matrix $L$ is defined as: $L=I-D^{-1/2}AD^{-1/2}$. It is a symmetric positive semi-definite matrix and has a complete set of eigenvectors $U$ formed by $\{u_s\}^{N-1}_{s=0}$, where $N$ is the number of vertices. The Laplacian of graph $\mathcal{G}$ can be diagonalized as $L=U\Lambda U^T$. Then we have graph Fourier transform $\hat{x}=U^T x$, which transforms the graph signal $x$ into spectral domain spanned by basis $U$.

Generalizing the convolution theorem into structured space on graph, convolution can be defined through decomposing a graph signal $\mathbf{s}\in \mathbb{R}^n$ on the spectral domain and then applying a spectral filter $g_\theta$ \cite{bruna2013spectral}. Naive implementation requires explicitly computing the Laplacian eigenvectors. To circumvent this problem, later study \cite{defferrard2016convolutional} approximated the spectral filter $g_{\theta}(\Lambda)$ with Chebyshev polynomials up to $K^{th}$ order, \emph{i.e.}, $g_{\theta}(\Lambda)\approx\sum_{k=0}^K \theta_k T_k(\Lambda)$, and then convolution of the graph signal can be formulated as:
\begin{equation}
g_{\theta}\star\mathbf{s}=\sum_{k=0}^K\theta_k T_k(L)\mathbf{s},
\end{equation}
where $T_k$ is the Chebyshev polynomials and $\{\theta_k\}$ is a vector of Chebyshev coefficients. In \cite{kipf2016semi}, the formulation is further simplified by limiting $K=1$, and approximating the largest eigenvalue of $L$ by $2$. In this way, the convolution becomes:
\begin{equation}
g_{\theta}\star\mathbf{s}=\theta\left(I+D^{-\frac{1}{2}}AD^{-\frac{1}{2}}\right)\mathbf{s},
\end{equation}
with $\theta$ being the only Chebyshev coefficient left. They further introduce a normalization trick:  
\begin{equation}
I+D^{-\frac{1}{2}}AD^{-\frac{1}{2}} \rightarrow \tilde{D}^{-\frac{1}{2}}\tilde{A}\tilde{D}^{-\frac{1}{2}}
\end{equation}
where $\tilde{A} = A + I$ and $\tilde{D}_{ii}=\sum_j\tilde{A}_{ij}$. Generalizing the convolution to a graph signal with $c$ channels, the layer-wise propagation rule in a multi-layer graph convolutional network (GCN) is given by \cite{kipf2016semi}:
\begin{equation}
H^{(l+1)}=\sigma\left(\tilde{D}^{-\frac{1}{2}}\tilde{A}\tilde{D}^{-\frac{1}{2}}H^{(l)}\Theta^{(l)}\right)
\label{forward}
\end{equation}
where $H^{(l)}$ is the vertices features of the $l$-th layer, $\Theta^{(l)}$ is the trainable weight matrix in layer $l$, and $\sigma$ is the non-linear activation function. 

The Eq~(\ref{forward}) provides the basis of performing convolution on graph-structured data, as adopted in \cite{zhang2018end,ying2018hierarchical}. For visual recognition tasks, in order to overcome the limited receptive field in current CNN architectures, some recent studies transform feature maps into region-based representation by projecting, and then perform graph reasoning with Eq~(\ref{forward}) to capture global relationships \cite{li2018beyond,liang2018symbolic,glore}.

\subsection{Graph Reasoning on Spatial Features}

Assuming that the propagation rule in Eq~(\ref{forward}) is applied on CNN features, \emph{i.e.}, $H^{(l)}=X^{(l)}\in\mathbb{R}^{H\times W\times C}$, the only difference between a GCN layer and a convolution layer is the graph Laplacian matrix $\tilde{L}=\tilde{D}^{-\frac{1}{2}}\tilde{A}\tilde{D}^{-\frac{1}{2}}$ applied on the left of $X^{(l)}$. In our study, we note that the original grid-based feature space can be deemed as a special case of data defined on a simple low-dimensional graph \cite{henaff2015deep}. Besides, the projecting process in current methods \cite{li2018beyond,liang2018symbolic,glore} actually achieves a similar purpose with the graph Laplacian matrix. They perform left multiplication on the input feature using a similarity matrix to have a global perception among all spatial locations. Therefore, we directly perform our graph reasoning in the original feature space. We save the projecting and re-projecting processes, and perform left matrix multiplication on the input feature only once. 

The Laplacian matrices in most current studies are data-independent parameters to learn. In order to better capture intra spatial structure, we propose an improved Laplacian $\tilde{L}$ that ensures the long-range context pattern to learn is dependent on the input features and not restricted as a specific one. It is formulated with the symmetric normalized form:
\begin{equation}
\tilde{L}=I-\tilde{D}^{-\frac{1}{2}}\tilde{A}\tilde{D}^{-\frac{1}{2}},
\label{eq5}
\end{equation}
where $\tilde{D}=diag(d_1,d_2,\dots,d_{n})$, $d_i=\sum_j \tilde{A}_{ij}$, and $\tilde{A}\in \mathbb{R}^{n\times n}$ is the data-dependent similarity matrix. We set $n=H\times W$, where $H\times W$ denotes the number of spatial locations of the input feature. 

For similarity matrix $\tilde{A}$, Euclidean distance can be used to estimate the graph structure as suggested in \cite{henaff2015deep}. We choose dot-product distance to calculate $A$, because dot product has a more friendly implementation in current deep learning platforms. The similarity between position $i$ and $j$ is expressed as:
\begin{equation}
\tilde{A}_{ij}=\phi(X)_i\tilde{\Lambda}(X)\phi(X)^T_j,
\label{eq6}
\end{equation}
where $\phi(X)\in \mathbb{R}^{HW\times M}$ is a linear embedding followed by ReLU$(\cdot)$ non-linearity, $M$ is the reduced dimension after transformation, and $\tilde{\Lambda}(X)\in \mathbb{R}^{M\times M}$ is a diagonal matrix that has position-agnostic attention on the inner product. It essentially learns a better distance metric for the similarity matrix $\tilde{A}$. Both $\phi(X)$ and $\tilde{\Lambda}(X)$ are data-dependent. Concretely, $\phi(X)$ is implemented as a $1\times1$ convolution, and $\tilde{\Lambda}(X)$ is implemented in a similar way as the channel-wise attention proposed in \cite{hu2018squeeze}. We calculate $\tilde{\Lambda}(X)$ as: 
\begin{equation}
\tilde{\Lambda}(X)=diag\left(\rho\left(\bar{X}\right)\right),
\label{eq7}
\end{equation}
where $\bar{X}\in\mathbb{R}^{1\times1\times C}$ is the feature after global pooling, and $\rho(\cdot)$ is another linear embedding with $1\times1$ convolution that reduce the dimension from $C$ to $M$. It is followed by the sigmoid function.

\begin{figure}
	\begin{center}
		\includegraphics[width=1\linewidth]{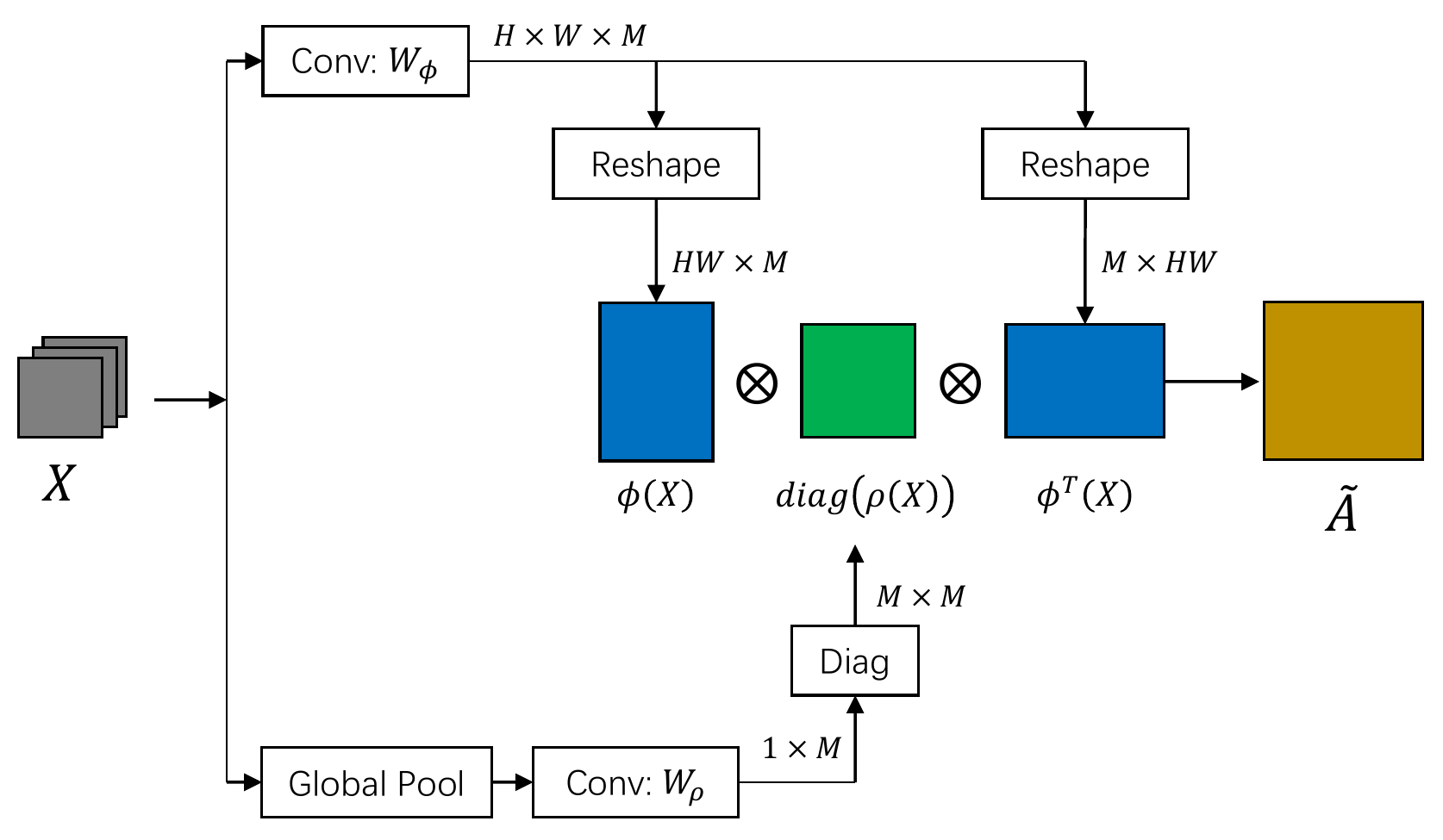}
	\end{center}
	\vspace{-2mm}
	\caption{The computation procedures of the similarity matrix $\tilde{A}$ from the input feature $X$.}
	\label{fig2}
	\vspace{-2mm}
\end{figure}

The computation procedures of $\tilde{A}$ is shown in Figure~\ref{fig2}, and we have its formulation as follows:
\begin{equation}
\tilde{A} = \phi \left(X;W_{\phi}\right) diag\left(\rho\left(\bar{X};W_{\rho}\right)\right) \phi \left(X;W_{\phi}\right)^T,
\end{equation}
where $W_{\phi}$ and $W_{\rho}$ are learnable parameters for the linear transformations. Because the degree matrix $\tilde{D}$ in Eq~(\ref{eq5}) has a function of normalization, we do not perform softmax on the similarity matrix $\tilde{A}$. Then we formulate the graph reasoning in our model as:
\begin{equation}
Y=\sigma\left(\tilde{L}X\Theta\right),
\label{eq9}
\end{equation}
where $X$ is the input feature, $\Theta$ is a trainable weight matrix, $\sigma$ is the ReLU activation function, and $Y$ is the output feature.

\subsection{Graph Reasoning on Spatial Pyramid}

Although graph reasoning is capable of capturing global context, we note that the same image contains multiple long-range contextual patterns. For examples, the finer representation may have more detailed long-range context, while the coarser representation provide more global dependencies. Since our graph reasoning module is directly performed in the original feature space, we organize the input feature as a spatial pyramid to extend the long-range contextual patterns that our method can capture. 

As shown in Figure~\ref{fig1}, graph reasonings are performed on each scale acquired by down-sampling, and then the output features are combined through up-sampling. It has a similar form with the feature pyramid network in \cite{lin2017feature}. But we implement our method on the final predicting feature, instead of the multi-scale features from the CNN backbone. Our graph reasoning on spatial pyramid can be expressed as follows:
\begin{equation}
\begin{split}
Y^{(s+1)} & ={\rm GR}(X^{(s+1)})+\Pi_{{\rm up}}(Y^{(s)}), \\
Y^{(0)} & ={\rm GR}(X^{(0)}), \\
X^{(s)} & =\Pi_{{\rm down}}(X^{(s+1)}), \\
\end{split}
\label{eq10}
\end{equation}
where ${\rm GR}$ denotes the graph reasoning with Eq~(\ref{eq9}), $s\ge0$ denotes the level of scales, and $\Pi_{{\rm up}},\Pi_{{\rm down}}$ represents the up-sampling and down-sampling operators, respectively. We implement $\Pi_{{\rm down}}$ using max-pooling with stride of $2$, and $\Pi_{{\rm up}}$ simply by bilinear interpolation. 

\subsection{Complexity Analysis}

In region-based graph reasoning studies \cite{li2018beyond,liang2018symbolic,glore}, they transform the grid-based CNN features into region-based vertices by projecting, which reduces the computational overhead for graph reasoning because the number of vertices is usually less than that of spatial locations. It seems that our method consumes more computation since we implement graph reasoning directly in the original feature space. Actually, we adopt an efficient computing strategy that successfully reduces the computational complexity of our method. We note that large computation is caused by the similarity matrix $\tilde{A}\in\mathbb{R}^{HW\times HW}$, therefore we do not explicitly calculate $\tilde{A}$. Concretely, we calculate the degree matrix $\tilde{D}$ in Eq~(\ref{eq5}) as follow:
\begin{equation}
\tilde{D}=diag\left(\tilde{A}\cdot\vec{1}\right)=diag\left(\phi\left(\tilde{\Lambda}\left(\phi^T\cdot\vec{1}\right)\right)\right)
\label{eq11}
\end{equation}
where $\vec{1}$ denotes an all-one vector in $\mathbb{R}^{HW}$. The brackets indicate the computation superiority. In this way, each step in Eq~(\ref{eq11}) is a multiplication with a vector, which effectively reduces the computational overhead. And then we calculate the left product of the Laplacian on the input feature as follows:
\begin{equation}
\begin{split}
\tilde{L}X & =X-\tilde{D}^{-\frac{1}{2}}\phi\tilde{\Lambda}\phi^T\tilde{D}^{-\frac{1}{2}}X \\
& = X-P\left(\tilde{\Lambda}\left(P^TX\right)\right) \\
\end{split}
\end{equation}
where $P$ is defined as $P=\tilde{D}^{-\frac{1}{2}}\phi$. Correspondingly, we calculate the terms in inner bracket first. In this way, we circumvent quadratic order of computation on the spatial locations $\mathcal{O}(H^2W^2)$. 

In our experiments, we set $C$ as $512$, and $M$ as $64$. Assuming that height $H$ and width $W$ of the input features are $97$, we calculate the computational and memory cost of our proposed layer, and compare with related methods in the same settings. As shown in Table~\ref{tab-overhead}, for our method on single-scale input, it has low computational cost. When we have spatial pyramid on $4$ scales, the computational and memory overheads do not show drastic increment. Therefore, our SpyGR layer does not introduce unbearable overhead in spite of its directly performing graph reasoning in the original feature space.

\begin{table}[t!]
	\renewcommand\arraystretch{1.1}
	\centering
	\setlength\tabcolsep{2mm}
	\begin{tabular}{l|cc}
		\hline
		Method 	& FLOPs (G) 	& Memory (M) 	\\ 
		\hline
		\hline
		Nonlocal~\cite{wang2018non}
		& 14.60	& 1072		\\
		
		$A^2$Net~\cite{chen20182}		
		& 3.11		& 110	\\ 		
		
		GloRe~\cite{glore}		
		& 3.11		& 103	\\ 
		SGR~\cite{liang2018symbolic}	
		& 6.24		& 118	\\
		
		DANet~\cite{fu2018dual}
		& 19.54	& 1114	\\		
		\hline
		\textbf{SpyGR} w/o pyramid
		& 3.11		& 120	\\ 
		\textbf{SpyGR}	
		& 4.12		& 164	\\ \hline
	\end{tabular}
	\caption{Overhead of different modules with input feature in $ \left[ 1 \times 512 \times 97 \times 97 \right] $. We show the complexity of our model on single-scale feature, and on a spatial pyramid with $4$ scales in the bottom two rows of the table.}
	\label{tab-overhead}
	\vspace{-3mm}
\end{table}

\section{Experiments}

\subsection{Datasets and Implementation Details}

To evaluate our proposed SpyGR layer, we carry out comprehensive experiments on the Cityscapes dataset~\cite{cityscapes}, the PASCAL Context dataset~\cite{pcontext} and the COCO Stuff dataset~\cite{coco}. We describe these datasets, together with implement details and loss function as follows.




\begin{table*}[t!]
	\begin{center}
		\setlength\tabcolsep{0.9mm}
		\small
		\begin{tabular}{l|c|ccccccccccccccccccc}
			\hline
			Method                                                  & mIoU & \rotatebox{90}{road} & \rotatebox{90}{sidewalk} & \rotatebox{90}{building} & \rotatebox{90}{wall} & \rotatebox{90}{fence} & \rotatebox{90}{pole} & \rotatebox{90}{traffic} \rotatebox{90}{light} & \rotatebox{90}{traffic} \rotatebox{90}{sign} & \rotatebox{90}{vegetation} & \rotatebox{90}{terrain} & \rotatebox{90}{sky}  & \rotatebox{90}{person} & \rotatebox{90}{rider} & \rotatebox{90}{car}  & \rotatebox{90}{truck} & \rotatebox{90}{bus}  & \rotatebox{90}{train} & \rotatebox{90}{motorcycle} & \rotatebox{90}{bicycle} \\ \hline \hline
			Deeplabv2~\cite{chen2018deeplab}  & 70.4 & 97.9 & 81.3     & 90.3     & 48.8 & 47.4  & 49.6 & 57.9          & 67.3         & 91.9       & 69.4    & 94.2 & 79.8   & 59.8  & 93.7 & 56.5  & 67.5 & 57.5  & 57.7       & 68.8    \\
			RefineNet~\cite{lin2017refinenet} & 73.6 & 98.2 & 83.3     & 91.3     & 47.8 & 50.4  & 56.1 & 66.9          & 71.3         & 92.3       & 70.3    & 94.8 & 80.9   & 63.3  & 94.5 & 64.6  & 76.1 & 64.3  & 62.2       & 70.0    \\
			DUC-HDC~\cite{duc-hdc}            & 77.6 & 98.5 & 85.5     & 92.8     & 58.6 & 55.5  & 65.0 & 73.5          & 77.9         & 93.3       & 72.0    & 95.2 & 84.8   & 68.5  & 95.4 & 70.9  & 78.8 & 68.7  & 65.9       & 73.8    \\
			SAC~\cite{sac}                    & 78.1 & \textbf{98.7} & 86.5     & 93.1     & 56.3 & 59.5  & 65.1 & 73.0          & 78.2         & 93.5       & \underline{72.6}    & 95.6 & 85.9   & 70.8  & 95.9 & 71.2  & 78.6 & 66.2  & 67.7       & 76.0    \\
			DepthSeg~\cite{depthseg}          & 78.2 & 98.5 & 85.4     & 92.5     & 54.4 & 60.9  & 60.2 & 72.3          & 76.8         & 93.1       & 71.6    & 94.8 & 85.2   & 69.0  & 95.7 & 70.1  & 86.5 & 75.7  & 68.3       & 75.5    \\
			PSPNet~\cite{zhao2017pyramid}     & 78.4 & -    & -        & -        & -    & -     & -    & -             & -            & -          & -       & -    & -      & -     & -    & -     & -    & -     & -          & -       \\
			AAF~\cite{adaptive}               & 79.1 & 98.5 & 85.6     & 93.0     & 53.8 & 59.0  & 65.9 & 75.0          & 78.4         & 93.7       & 72.4    & 95.6 & 86.4   & 70.5  & 95.9 & 73.9  & 82.7 & 76.9  & 68.7       & 76.4    \\
			DFN~\cite{dfn}                    & 79.3 & -    & -        & -        & -    & -     & -    & -             & -            & -          & -       & -    & -      & -     & -    & -     & -    & -     & -          & -       \\
			PSANet~\cite{zhao2018psanet}      & 80.1 & -    & -        & -        & -    & -     & -    & -             & -            & -          & -       & -    & -      & -     & -    & -     & -    & -     & -          & -       \\
			DenseASPP~\cite{denseaspp}        & 80.6 & \textbf{98.7} & \textbf{87.1}     & 93.4     & \textbf{60.7} & 62.7  & 65.6 & 74.6          & 78.5         & 93.6       & 72.5    & 95.4 & 86.2   & 71.9  & 96.0 & \textbf{78.0}  & \textbf{90.3} & 80.7  & 69.7       & 76.8    \\
			GloRe~\cite{glore}                & 80.9 & -    & -        & -        & -    & -     & -    & -             & -            & -          & -       & -    & -      & -     & -    & -     & -    & -     & -          & -       \\
			DANet~\cite{fu2018dual}           & \underline{81.5} & 98.6 & 86.1     & \underline{93.5}     & 56.1 & \textbf{63.3}  & \underline{69.7} & \underline{77.3}          & \underline{81.3}         & \textbf{93.9}       & \textbf{72.9}    & \textbf{95.7} & \underline{87.3}   & \underline{72.9}  & \textbf{96.2} & \underline{76.8}  & \underline{89.4} & \textbf{86.5}  & \textbf{72.2}       & \underline{78.2}    \\
			\textbf{SpyGR}                    & \textbf{81.6} & \textbf{98.7} & \underline{86.9}     & \textbf{93.6}     & \underline{57.6} & \underline{62.8}  & \textbf{70.3} & \textbf{78.7}          & \textbf{81.7}         & \underline{93.8}       & 72.4    & \underline{95.6} & \textbf{88.1}   & \textbf{74.5}  & \textbf{96.2} & 73.6  & 88.8 & \underline{86.3}  & \underline{72.1}       & \textbf{79.2}    \\ \hline
		\end{tabular}
	\end{center}
	\vspace{-3mm}
	\caption{Per-class results on Cityscapes testing set. Best results are marked in bold and the second best results are underlined.  It is shown that SpyGR achieves the highest performance and has superiority in most categories.}
	\label{tab-city}
	\vspace{-2mm}
\end{table*}

\noindent\textbf{Implement Details}. We use ResNet~\cite{he2016deep}~(pretrained on ImageNet~\cite{deng2009imagenet}) as our backbone. We use a $3 \times 3$ convolution to reduce the channel number from 2048 to 512, and then stack SpyGR layer upon it. We set $M$ as 64 in all our experiments. Following prior works~\cite{zhao2017pyramid,chen2018deeplab,chen2017rethinking}, we employ a polynomial learning rate policy where the initial learning rate is multiplied by $ \left( 1 - iter /\ total\_iter \right)^{0.9} $ after each iteration. Momentum and weight decay coefficients are set to 0.9 and 0.0001 respectively, and the base learning rate is set to 0.009 for all datasets. For data augmentation, we apply the common scale, cropping and  flipping strategies to augment the training data. Input size is set as $769 \times 769$ for Cityscapes, and $513 \times 513$ for others. The synchronized batch normalization is adopted in all experiments, together with the multi-grid~\cite{chen2017rethinking} scheme. For evaluation, we use the Mean IoU metric as a common choice. We downsample for three times and have four levels in our pyramid. 

\noindent\textbf{Loss Function}. We employ the standard cross entropy loss on both final output of our model, and the intermediate feature map output from res4b22. We set the weight over the final loss as 1 and the auxiliary loss as 0.4, following the settings in PSPNet~\cite{zhao2017pyramid}.

\subsection{Results on Cityscapes}

We first compare our method with existing methods on the Cityscapes test set. To fairly compare with others, we train our SpyGR upon ResNet-101 with output stride as 8. Note that we only train on fine annotated data. We adopt the OHEM scheme~\cite{ohem} for final loss, and train the model for 80K iterations, with mini-batch size set as 8. For testing, we adopt multi-scale~(0.75, 1.0, 1.25, 1.5, 1.75, 2.0) inference and flipping, and then submit the predictions to official evaluation server. Results are shown in Table~\ref{tab-city}. We can see that SpyGR shows superiority in most categories. SpyGR outperforms GloRe~\cite{glore}, the latest graph convolutional networks~(GCN) based model, by 0.7 in mIoU. Moreover, SpyGR even outperforms DANet, a recently proposed self-attention based model, whose computation overhead and memory requirements are much higher than our proposed methods, as shown in Table~\ref{tab-overhead}.

\subsection{Comparisons with DeepLabV3}

DeepLabV3 \cite{chen2017rethinking} and DeepLabV3+ \cite{chen2018encoder} report their results on Cityscapes by training on the \emph{fine+coarse} set. In order to show the effectiveness of our proposed methods over them, we conduct detailed comparisons on both Cityscapes and PASCAL VOC. As shown in Table~\ref{deeplab}, SpyGR consistently has at least 1 mIoU gains over DeepLabV3. The advantages of SpyGR over DeepLabV3+ are more significant on PASCAL VOC than Cityscapes. 

\subsection{Results on COCO Stuff}

For the COCO Stuff dataset, we train SpyGR with output stride of 8, and mini-batch size of 12. We train for 30K iterations on the COCO Stuff training set, around 40 epochs, which is much shorter than DANet's 240 epochs. Multi-scale input and flipping are used for testing. The comparison  on the COCO Stuff dataset is shown in Table~\ref{tab-coco}. Similar to the other two datasets, our SpyGR also outperforms other methods performance on the COCO Stuff dataset. It has a comparable result with DANet, but shows a significant superiority over SGR. 

\subsection{Results on PASCAL Context}
We carry out experiments on the PASCAL Context dataset to further evaluate the validity of our proposed SpyGR. We train our model with mini-batch size of 16 and output stride of 16, and inference with output stride of 8. To make SpyGR operated with the same stride during both training and inference phase, we upsample C5 from ResNet-101, and concatenate it with C3, which has an output stride of 8. A $3 \times 3$ convolution is appended over the concatenation of C3 and C5, and then we add our SpyGR layer. We optimize the whole network on training set of PASCAL Context for 15K iterations, around 48 epochs. As a comparison, DANet trains for 240 epoch, around 5 times of us. For evaluation on test set, we adopt the multi-scale and flipping augmentations. We show the experimental results of PASCAL Context in Table~\ref{tab-pcontext}. It is shown that even SpyGR with ResNet-50 as backbone achieves comparable performance with SGR on ResNet-101, and outperforms MSCI~\cite{msci} on ResNet-152. Furthermore, SpyGR on ResNet-101 gains higher performance than SGR+, even though SGR+ is pre-trained on the COCO Stuff dataset. And once again, SpyGR outperforms DANet by a small margin, but with much less computational overhead and memory cost, and a significantly shorter training scheduler.

\begin{table}[t!]
	\renewcommand\arraystretch{1.1}
	\setlength\tabcolsep{1.1mm}
	\begin{center}
		\begin{tabular}{c|c|c|c|c|c|c}
			\hline
			\multicolumn{1}{l|}{\multirow{3}{*}{Methods}} & \multicolumn{3}{c|}{Cityscapes}               & \multicolumn{3}{c}{PASCAL VOC}                 \\ \cline{2-7} 
			\multicolumn{1}{l|}{}                         & \multicolumn{2}{c|}{Val}      & Test          & \multicolumn{2}{c|}{Val}        & Test          \\ \cline{2-7} 
			\multicolumn{1}{l|}{}                         & SS            & MS            & +Coarse       & SS             & MS             & Finetune      \\ \hline \hline
			DeepLabV3                                     & 78.3          & 79.3          & 81.3          & 78.5          & 79.8          & -             \\
			DeepLabV3+                                    & 79.6          & 80.2          & 82.1          & 79.4          & 80.4          & 83.3          \\
			\hline
			\textbf{SpyGR}                                         & \textbf{79.9} & \textbf{80.5} & \textbf{82.3} & \textbf{80.2} & \textbf{81.2} & \textbf{84.2} \\ \hline
		\end{tabular}%
	\end{center}
	\vspace{-3mm}
	\caption{Comparisons with DeepLabV3. \textbf{SS} means single scale, \textbf{MS} denotes multi-scale. \textbf{+Coarse} means training on \textit{fine+coarse} set. \textbf{Finetune} means finetuning on the \emph{trainval} set. To be fair, all results of compared methods are tested on their newest implementations.}
	\label{deeplab}
\end{table}

\begin{table}[t!]
	\begin{center}
		\setlength\tabcolsep{5mm}
		\begin{tabular}{l|c|c}
			\hline
			Method         					& Backbone 		& mIoU~(\%) \\ \hline \hline
			RefineNet~\cite{lin2017refinenet}& ResNet-101    & 33.6 \\
			CCL~\cite{ccl}					& ResNet-101    & 35.7 \\
			DANet~\cite{fu2018dual}  		& ResNet-50     & 37.2 \\
			DSSPN~\cite{dsspn}          	& ResNet-101    & 37.3 \\
			\textbf{SpyGR}					& ResNet-50     & \underline{37.5} \\
			SGR~\cite{liang2018symbolic}	& ResNet-101    & 39.1 \\
			DANet~\cite{fu2018dual}        	& ResNet-101    & 39.7 \\
			\textbf{SpyGR}         			& ResNet-101    & \textbf{39.9} \\ \hline
		\end{tabular}
	\end{center}
	\vspace{-3mm}
	\caption{The comparison on the COCO Stuff test set.}
	\label{tab-coco}
	\vspace{-2mm}
\end{table}

\subsection{Ablation Studies}

We conduct ablation studies to explore how does each part of SpyGR contribute to the performance gain. We carry out all ablation experiments on Cityscapes over ResNet-50. For inference, we only use single-scale input image. The comparisons are listed in Table~\ref{tab-ablation}. We analyze each part of SpyGR as follow.

\noindent\textbf{Simplest GCN.} We consider the case without the attentional diagonal matrix. The similarity matrix $\tilde{A}$ reduces to:
\begin{equation}
\tilde{A} = \phi \left(X, W_{\phi}\right) \phi \left(X, W_{\phi}\right)^T.
\end{equation}
Removing the identity in Laplacian, the propagation rule of graph reasoning in Eq~(\ref{eq9}) now becomes as follow:
\begin{equation}
Y = \sigma \left( \tilde{D}^{-\frac{1}{2}} \tilde{A} \tilde{D}^{-\frac{1}{2}} X \Theta \right).
\end{equation}
The simplest GCN brings an increase of 1.64 in mIoU.

\begin{table}[t!]
	\begin{center}
		\setlength\tabcolsep{5.6mm}
		\begin{tabular}{l|c|c}
			\hline
			Method      					& Backbone 		  & mIoU~(\%) \\ \hline \hline
			PSPNet~\cite{zhao2017pyramid}	& ResNet-101      & 47.8 \\
			DANet~\cite{fu2018dual}   		& ResNet-50       & 50.1 \\
			MSCI~\cite{msci}     			& ResNet-152      & 50.3 \\
			\textbf{SpyGR}      			& ResNet-50       & \underline{50.3} \\
			SGR~\cite{liang2018symbolic}	& ResNet-101      & 50.8 \\
			CCL~\cite{ccl}       			& ResNet-101      & 51.6 \\
			EncNet~\cite{zhang2018context} 	& ResNet-101      & 51.7 \\
			SGR+~\cite{liang2018symbolic}	& ResNet-101      & 52.5 \\
			DANet~\cite{fu2018dual}      	& ResNet-101      & 52.6 \\
			\textbf{SpyGR}      			& ResNet-101      & \textbf{52.8} \\ \hline
		\end{tabular}
	\end{center}
	\vspace{-3mm}
	\caption{The comparison on the test set of PASCAL Context. `+' means pretrained on COCO Stuff.}
	\label{tab-pcontext}
\end{table}

\begin{table}[t!]
	\renewcommand\arraystretch{1.25}
	\begin{center}
		\setlength\tabcolsep{1.8mm}
		\begin{tabular}{cccccc|c}
			\hline
			FCN	& GCN	& $\tilde{\Lambda}$ & $\tilde{\Lambda}(X)$ & Identity& Pyramid & mIoU	\\ \hline \hline
			\ding{51}	& -			& -  				& - 		& - 		& - 		& 76.34 \\
			\ding{51}	& \ding{51}	& - 	 			& -  		& - 		& - 		& 77.98 \\
			\ding{51}	& \ding{51}	& \ding{51}			& -  		& - 		& -  		& 78.58 \\
			\ding{51}	& \ding{51}	& \ding{51}			& \ding{51}	& - 		& - 		& 79.05	\\
			\ding{51}	& \ding{51}	& \ding{51}			& \ding{51}	& \ding{51}	& -  		& 79.42 \\
			\ding{51}	& \ding{51}	& \ding{51}			& \ding{51}	& \ding{51}	& \ding{51}	& 79.93	\\ \hline
		\end{tabular}
	\end{center}
	\vspace{-3mm}
	\caption{Ablation experiments on the Cityscapes dataset.}
	\label{tab-ablation}
\end{table}

\begin{figure*}[th!]
	\centering
	\begin{subfigure}{0.16\textwidth}
		\includegraphics[width=\textwidth]{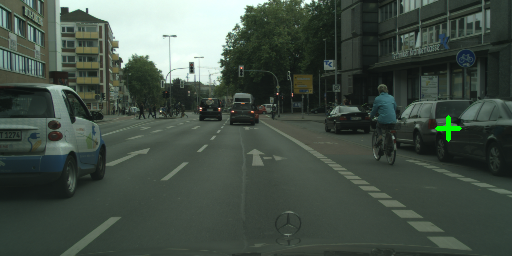}
	\end{subfigure}
	\begin{subfigure}{0.16\textwidth}
		\includegraphics[width=\textwidth]{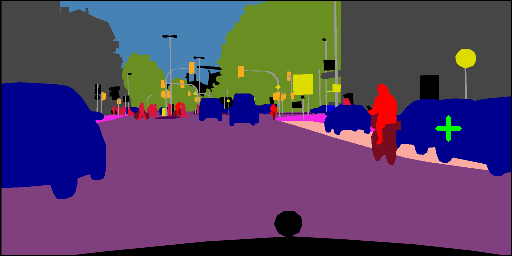}
	\end{subfigure}
	\begin{subfigure}{0.16\textwidth}
		\includegraphics[width=\textwidth]{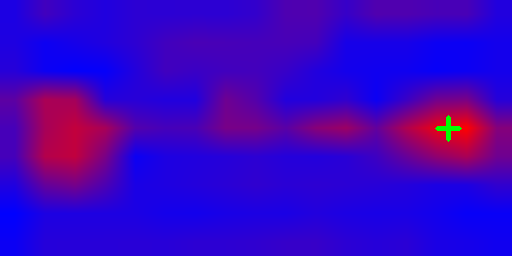}
	\end{subfigure}
	\begin{subfigure}{0.16\textwidth}
		\includegraphics[width=\textwidth]{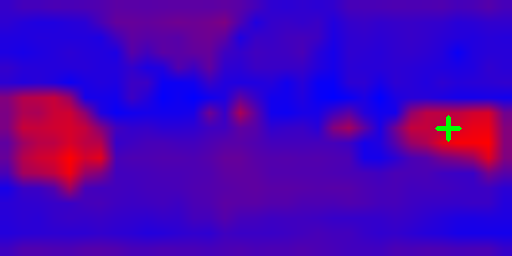}
	\end{subfigure}
	\begin{subfigure}{0.16\textwidth}
		\includegraphics[width=\textwidth]{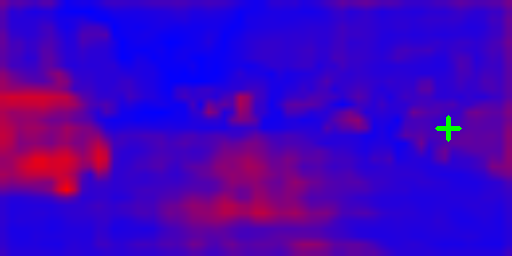}
	\end{subfigure}
	\begin{subfigure}{0.16\textwidth}
		\includegraphics[width=\textwidth]{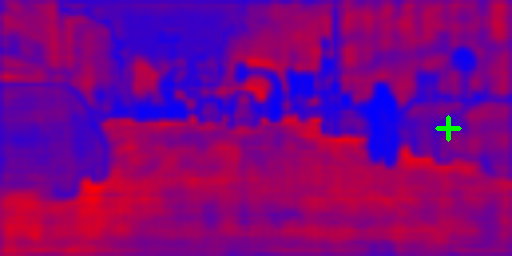}
	\end{subfigure}

	\vspace{0.5mm}
	
	\begin{subfigure}{0.16\textwidth}
		\includegraphics[width=\textwidth]{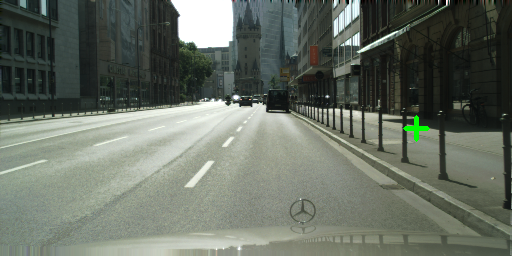}
	\end{subfigure}
	\begin{subfigure}{0.16\textwidth}
		\includegraphics[width=\textwidth]{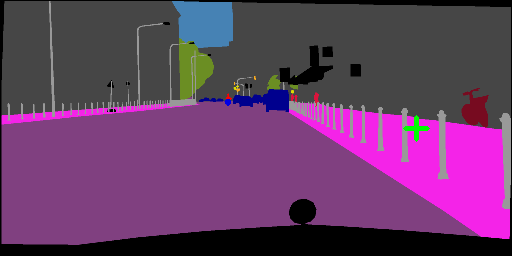}
	\end{subfigure}
	\begin{subfigure}{0.16\textwidth}
		\includegraphics[width=\textwidth]{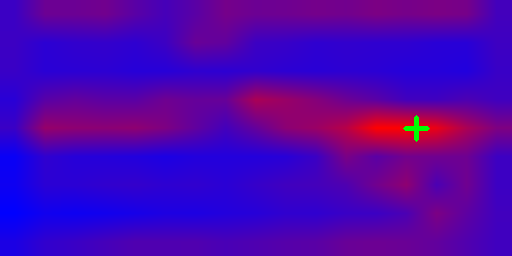}
	\end{subfigure}
	\begin{subfigure}{0.16\textwidth}
		\includegraphics[width=\textwidth]{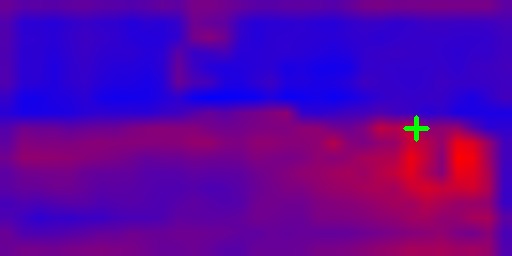}
	\end{subfigure}
	\begin{subfigure}{0.16\textwidth}
		\includegraphics[width=\textwidth]{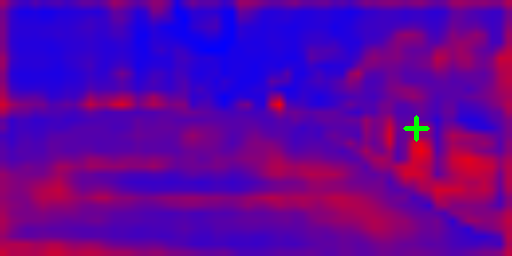}
	\end{subfigure}
	\begin{subfigure}{0.16\textwidth}
		\includegraphics[width=\textwidth]{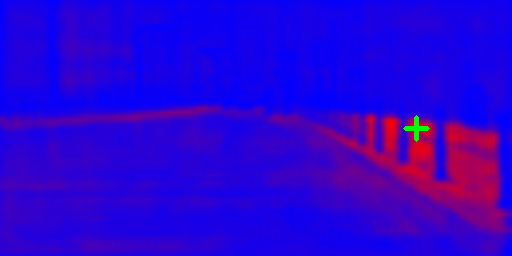}
	\end{subfigure}
	
	\vspace{0.5mm}
	
	\begin{subfigure}{0.16\textwidth}
		\includegraphics[width=\textwidth]{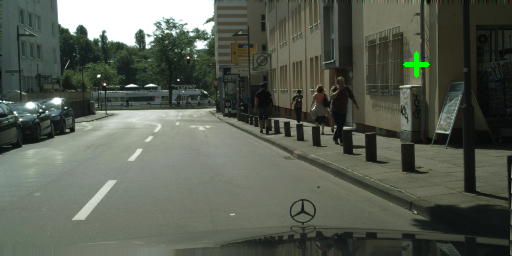}
	\end{subfigure}
	\begin{subfigure}{0.16\textwidth}
		\includegraphics[width=\textwidth]{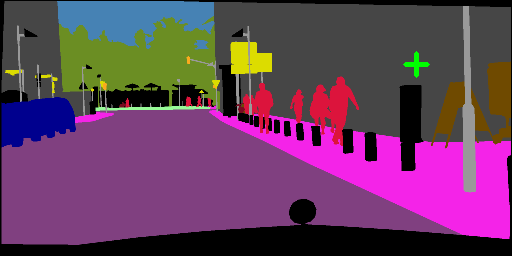}
	\end{subfigure}
	\begin{subfigure}{0.16\textwidth}
		\includegraphics[width=\textwidth]{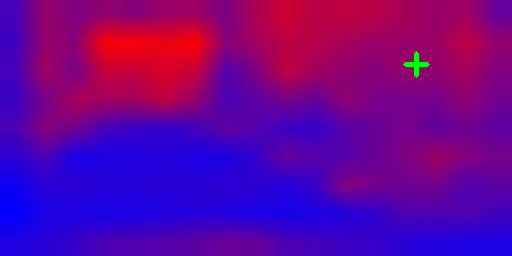}
	\end{subfigure}
	\begin{subfigure}{0.16\textwidth}
		\includegraphics[width=\textwidth]{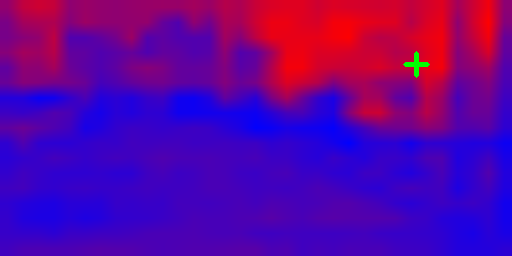}
	\end{subfigure}
	\begin{subfigure}{0.16\textwidth}
		\includegraphics[width=\textwidth]{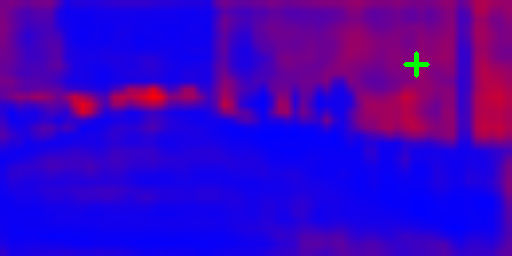}
	\end{subfigure}
	\begin{subfigure}{0.16\textwidth}
		\includegraphics[width=\textwidth]{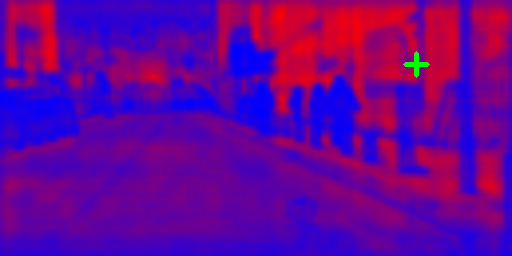}
	\end{subfigure}

	\vspace{-3mm}
	\caption{Visualization of the similarity matrix $\tilde{A}_i$ of a randomly sampled location $i$ marked in green cross. The left two columns are input images and ground truths respectively. The similarity matrix of different scales in the pyramid are re-scaled in the same size and shown in the right four columns from the coarsest to the finest (from left to right). Multiple long-range contextual patterns are captured in different scales, and aggregated in the finest level. Zoom in to have a better view.}
	\label{fig-Z}
\end{figure*}

\begin{figure*}[t!]
	\centering
	\begin{subfigure}{0.16\textwidth}
		\includegraphics[width=\textwidth]{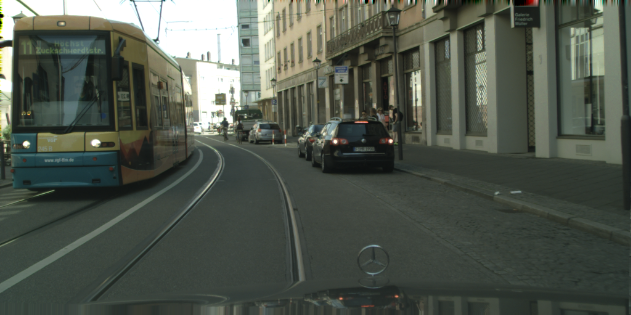}
	\end{subfigure}
	\begin{subfigure}{0.16\textwidth}
		\includegraphics[width=\textwidth]{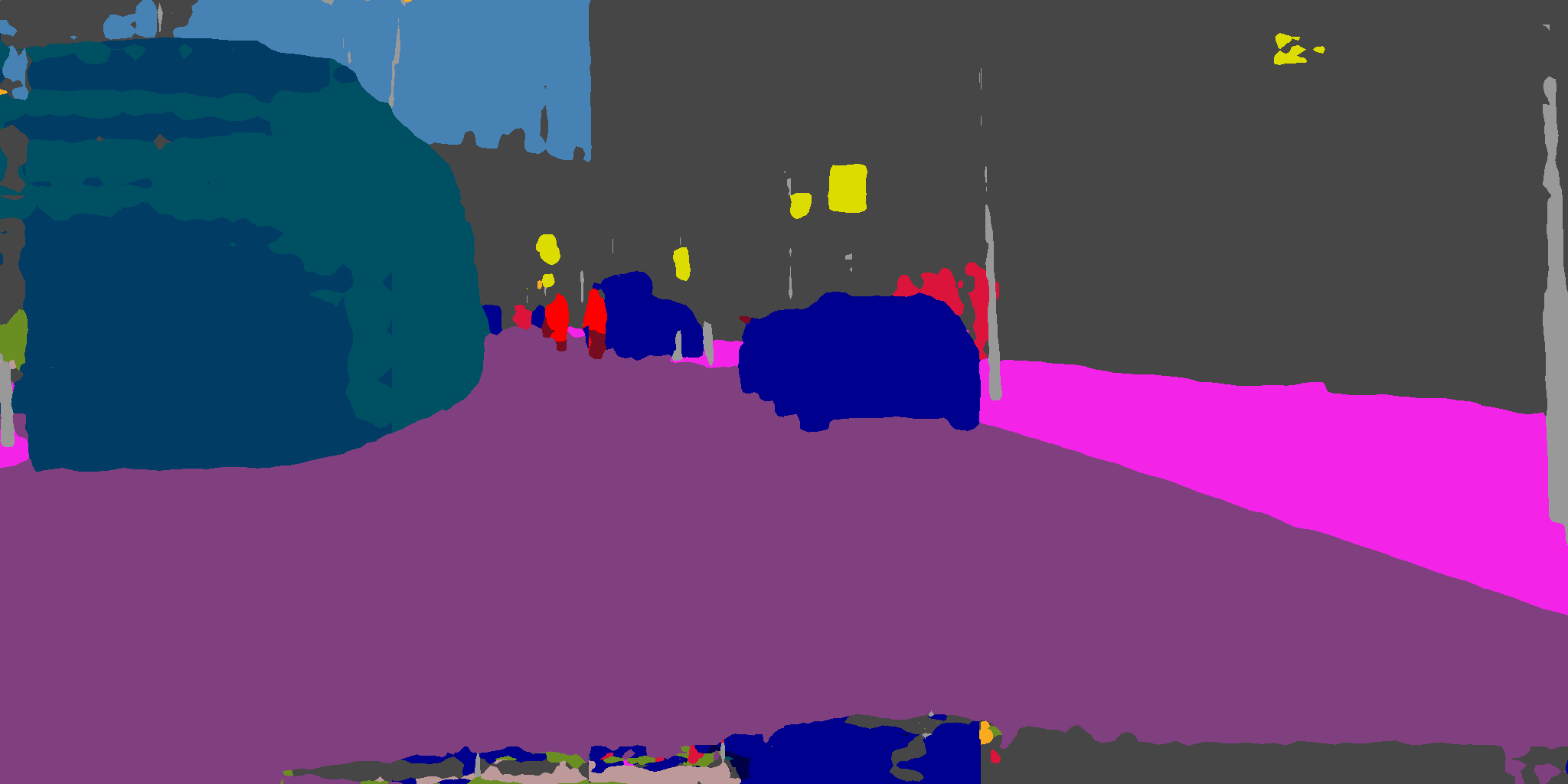}
	\end{subfigure}
	\begin{subfigure}{0.16\textwidth}
		\includegraphics[width=\textwidth]{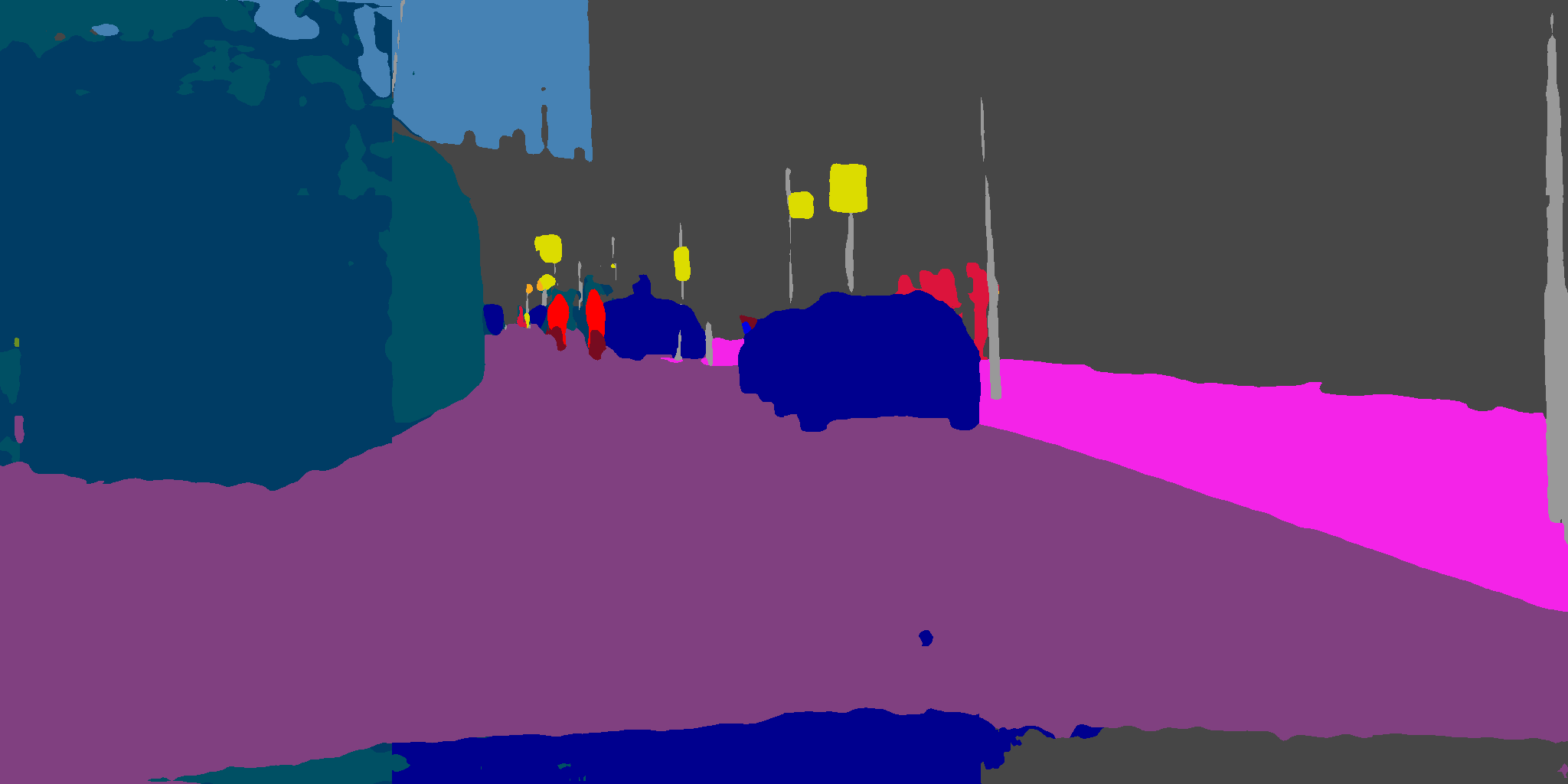}
	\end{subfigure}
	\begin{subfigure}{0.16\textwidth}
		\includegraphics[width=\textwidth]{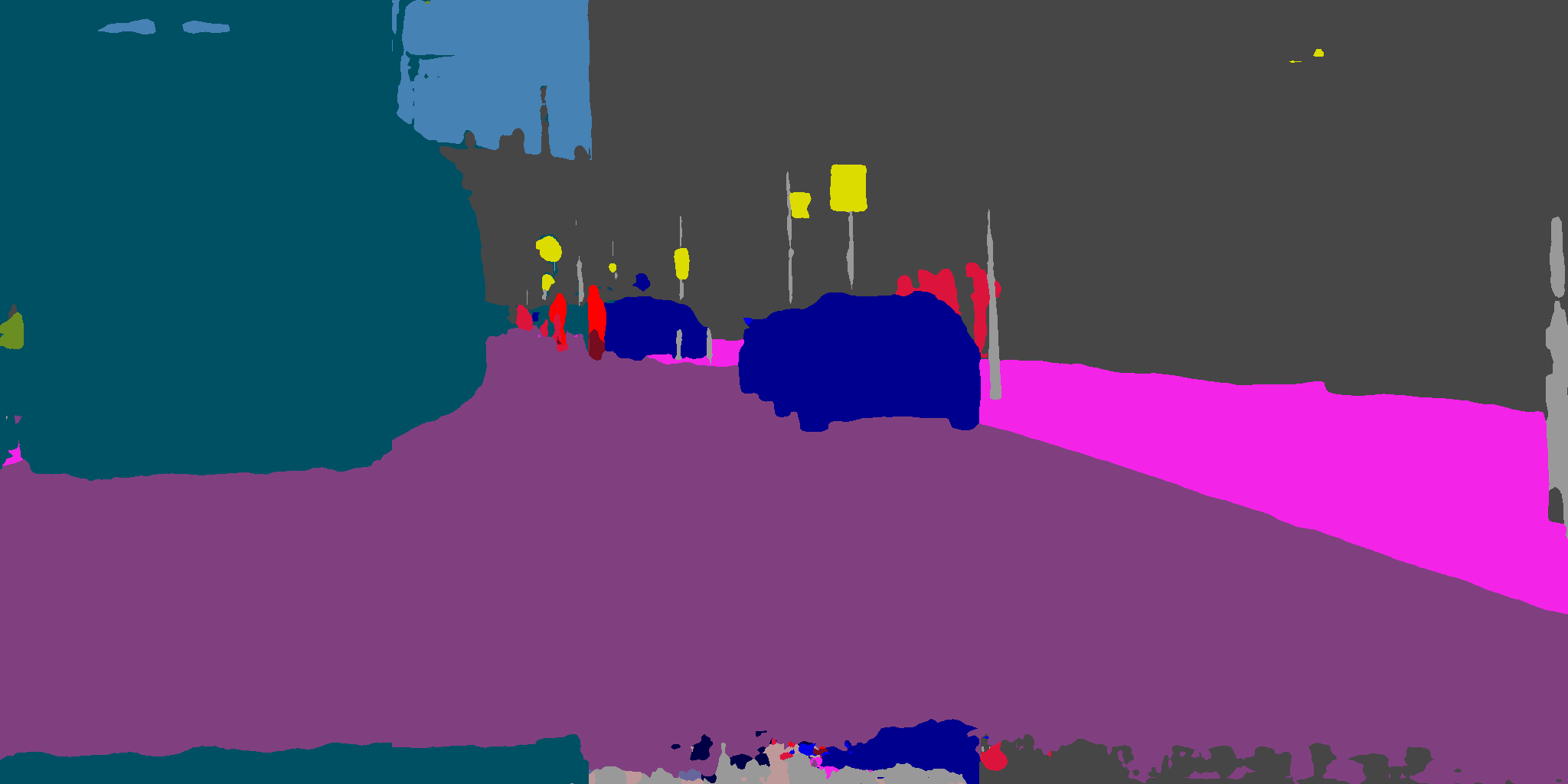}
	\end{subfigure}
	\begin{subfigure}{0.16\textwidth}
		\includegraphics[width=\textwidth]{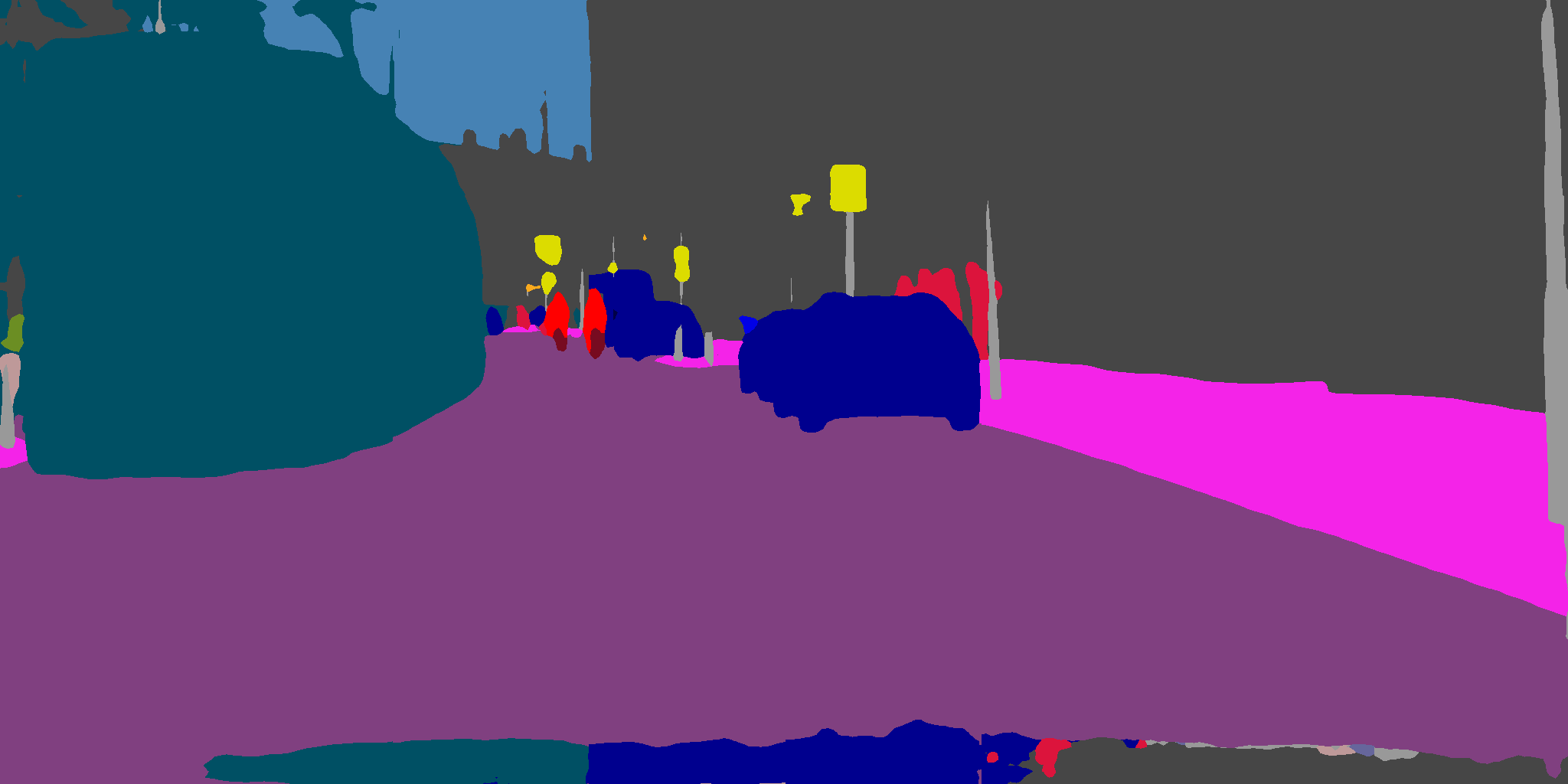}
	\end{subfigure}
	\begin{subfigure}{0.16\textwidth}
		\includegraphics[width=\textwidth]{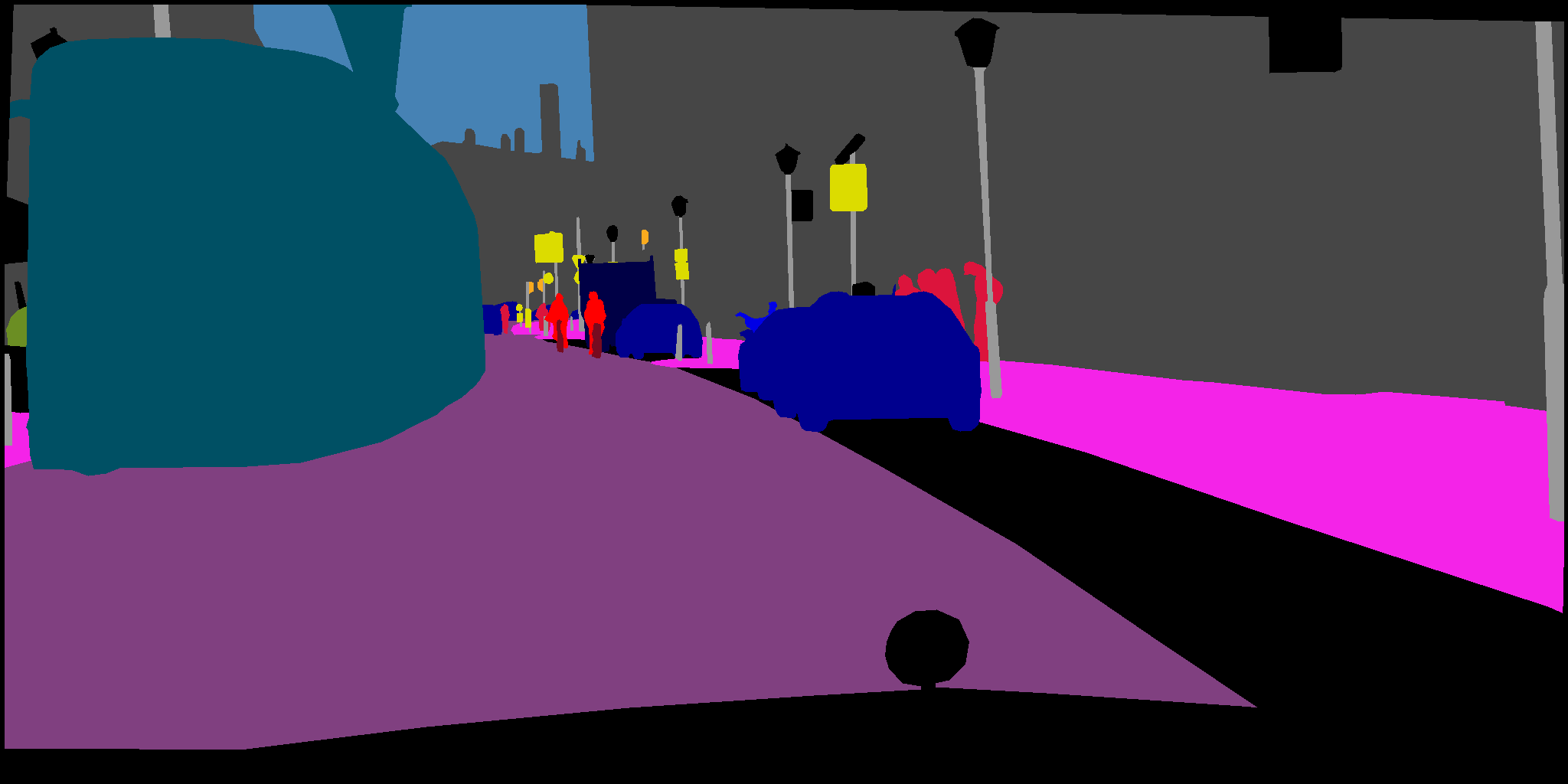}
	\end{subfigure}
	
	\vspace{0.5mm}
	
	\begin{subfigure}{0.16\textwidth}
		\includegraphics[width=\textwidth]{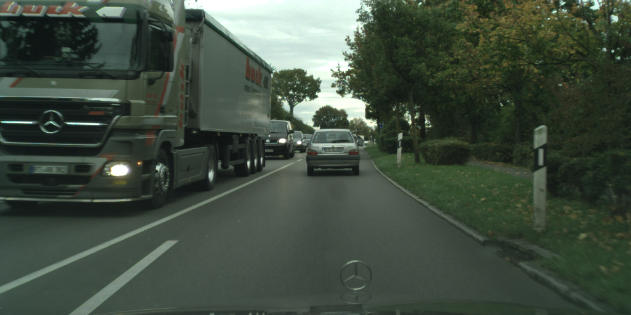}
	\end{subfigure}
	\begin{subfigure}{0.16\textwidth}
		\includegraphics[width=\textwidth]{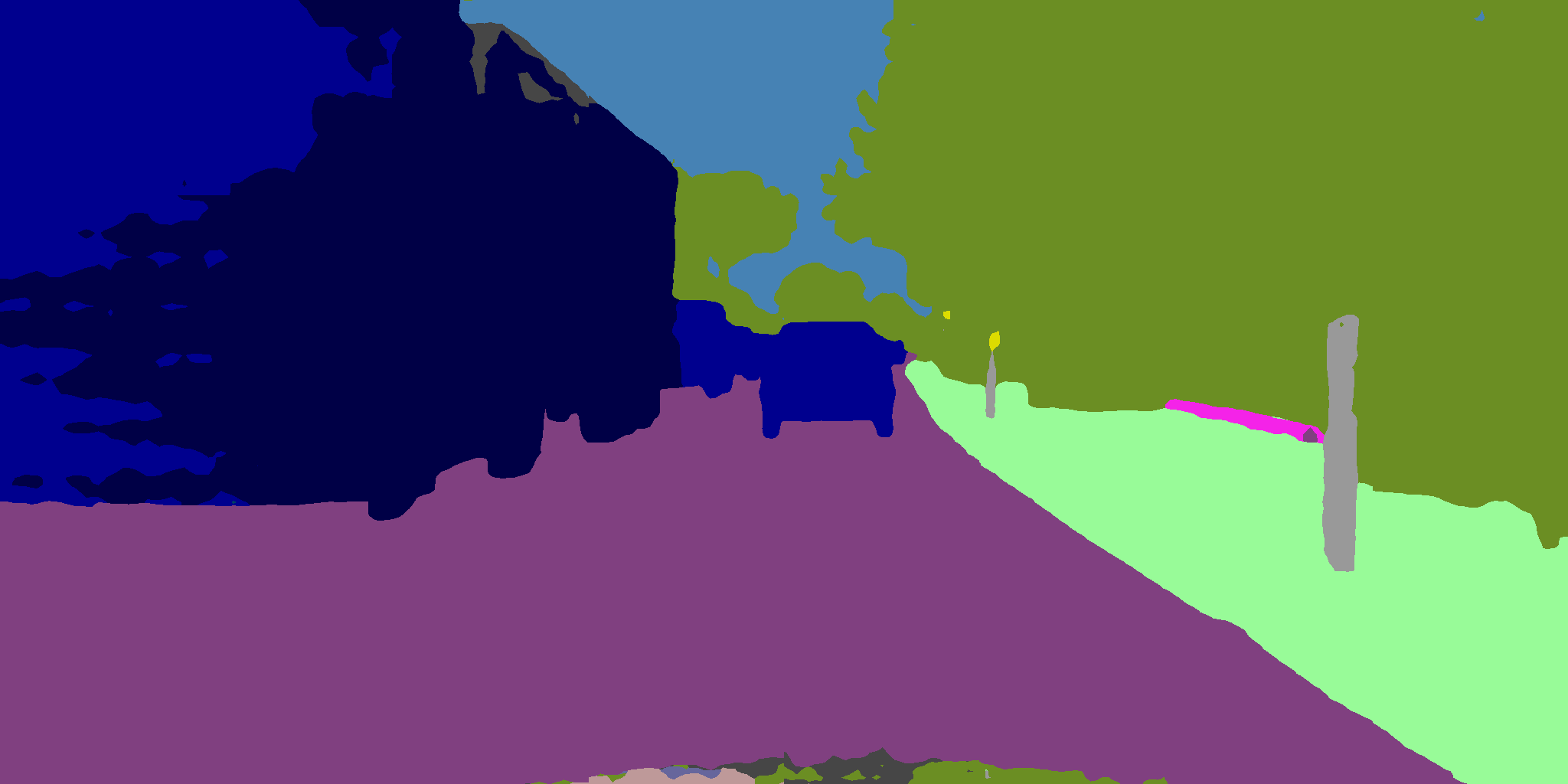}
	\end{subfigure}
	\begin{subfigure}{0.16\textwidth}
		\includegraphics[width=\textwidth]{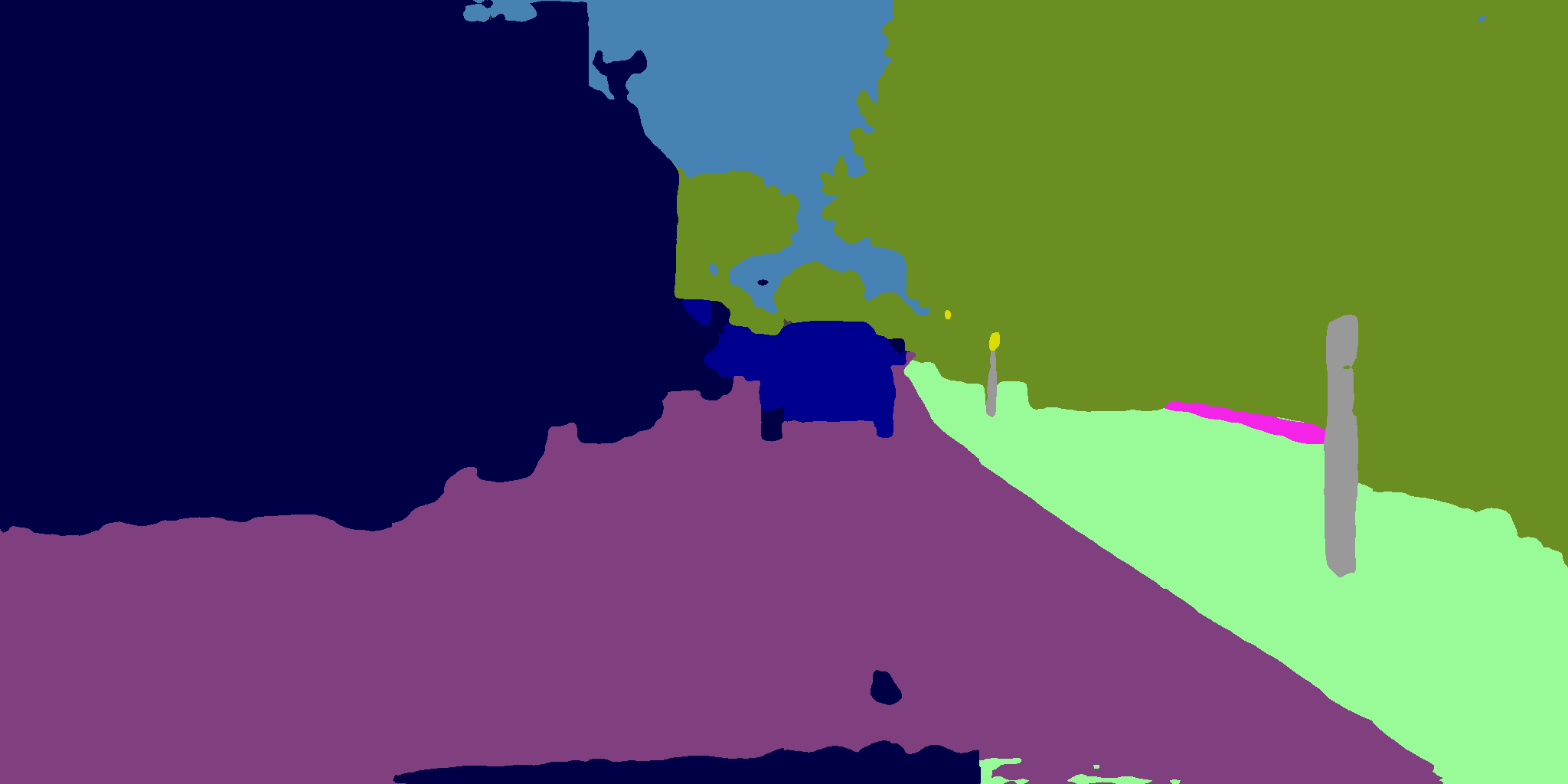}
	\end{subfigure}
	\begin{subfigure}{0.16\textwidth}
		\includegraphics[width=\textwidth]{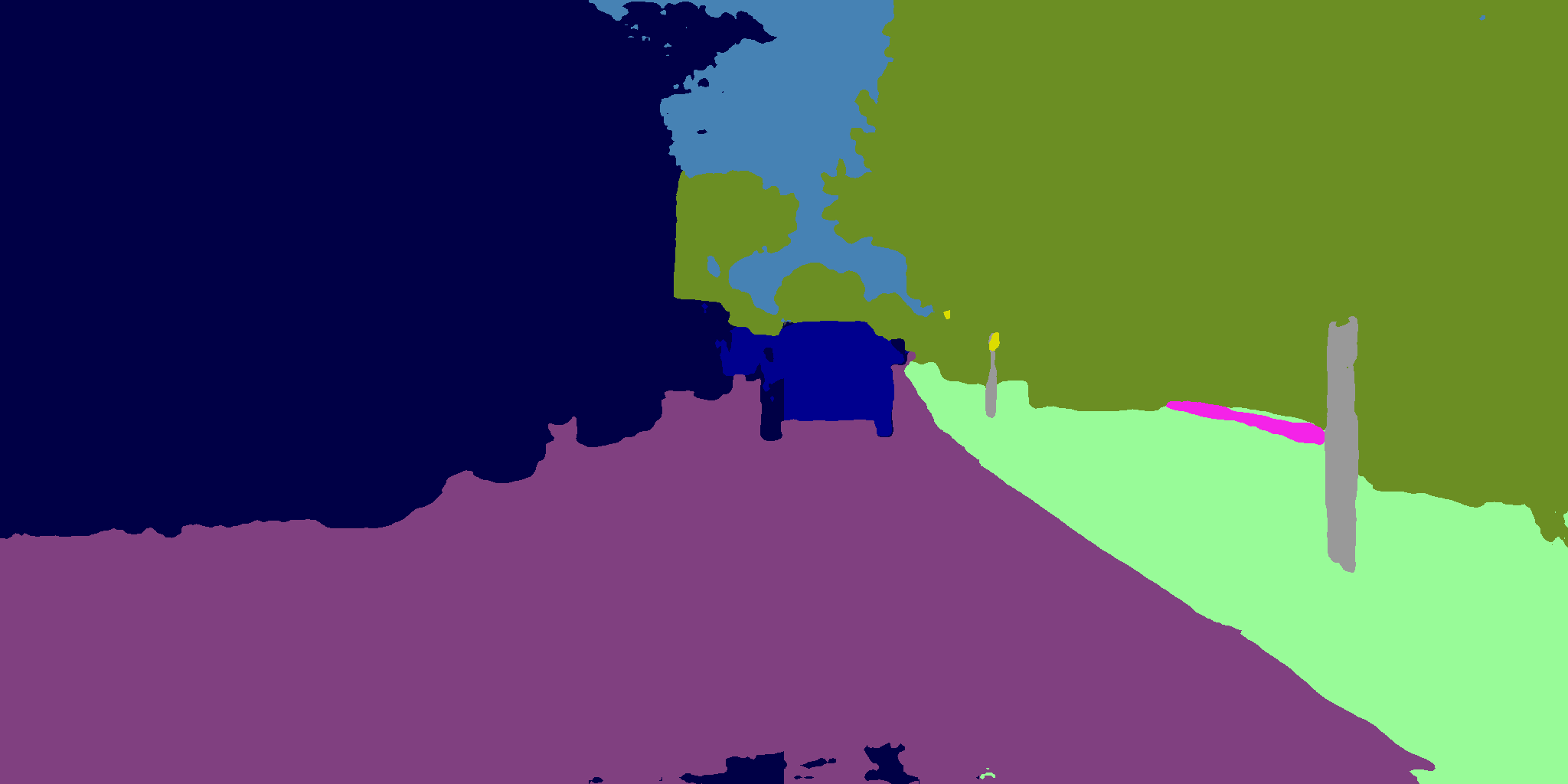}
	\end{subfigure}
	\begin{subfigure}{0.16\textwidth}
		\includegraphics[width=\textwidth]{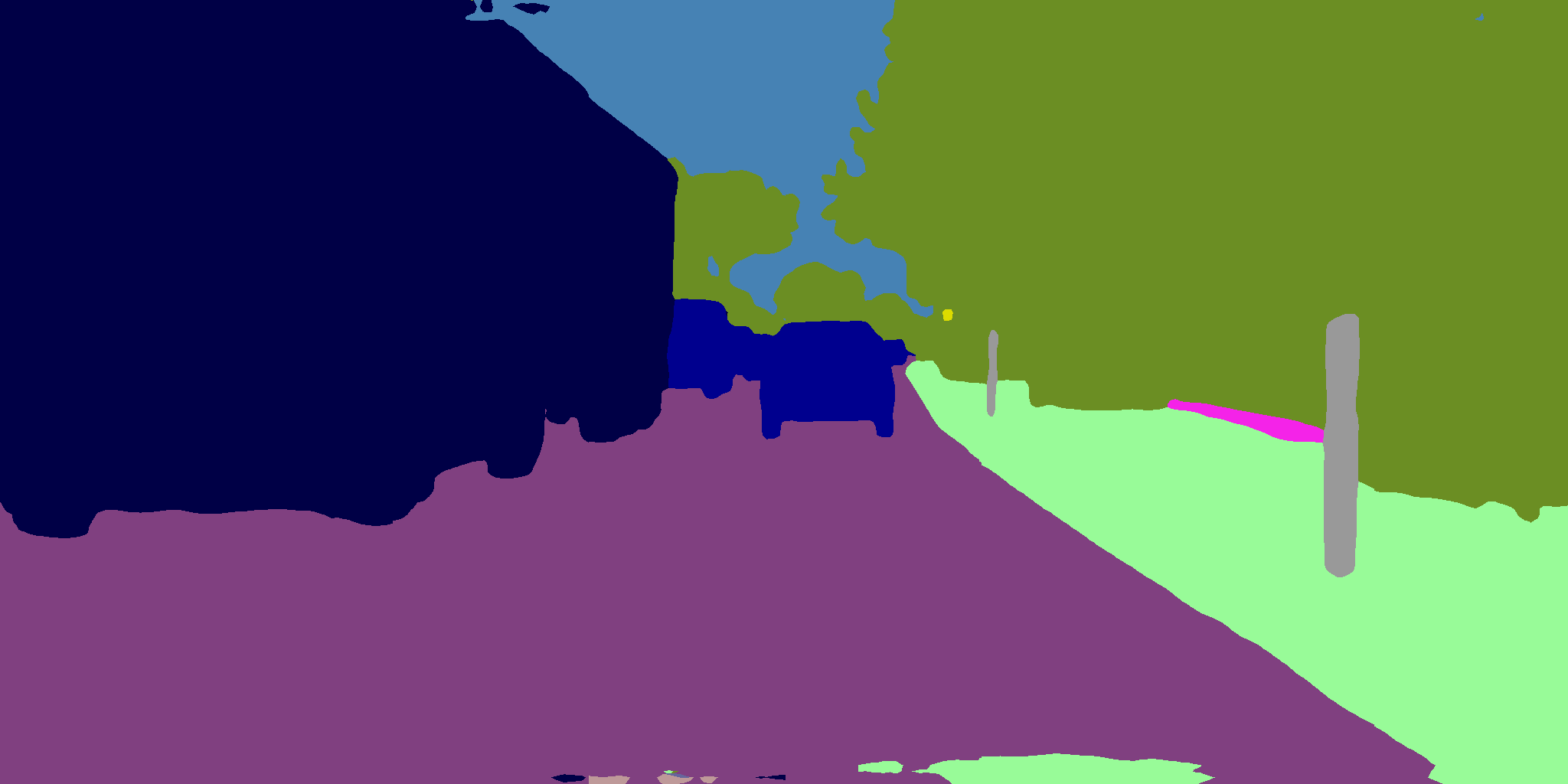}
	\end{subfigure}
	\begin{subfigure}{0.16\textwidth}
		\includegraphics[width=\textwidth]{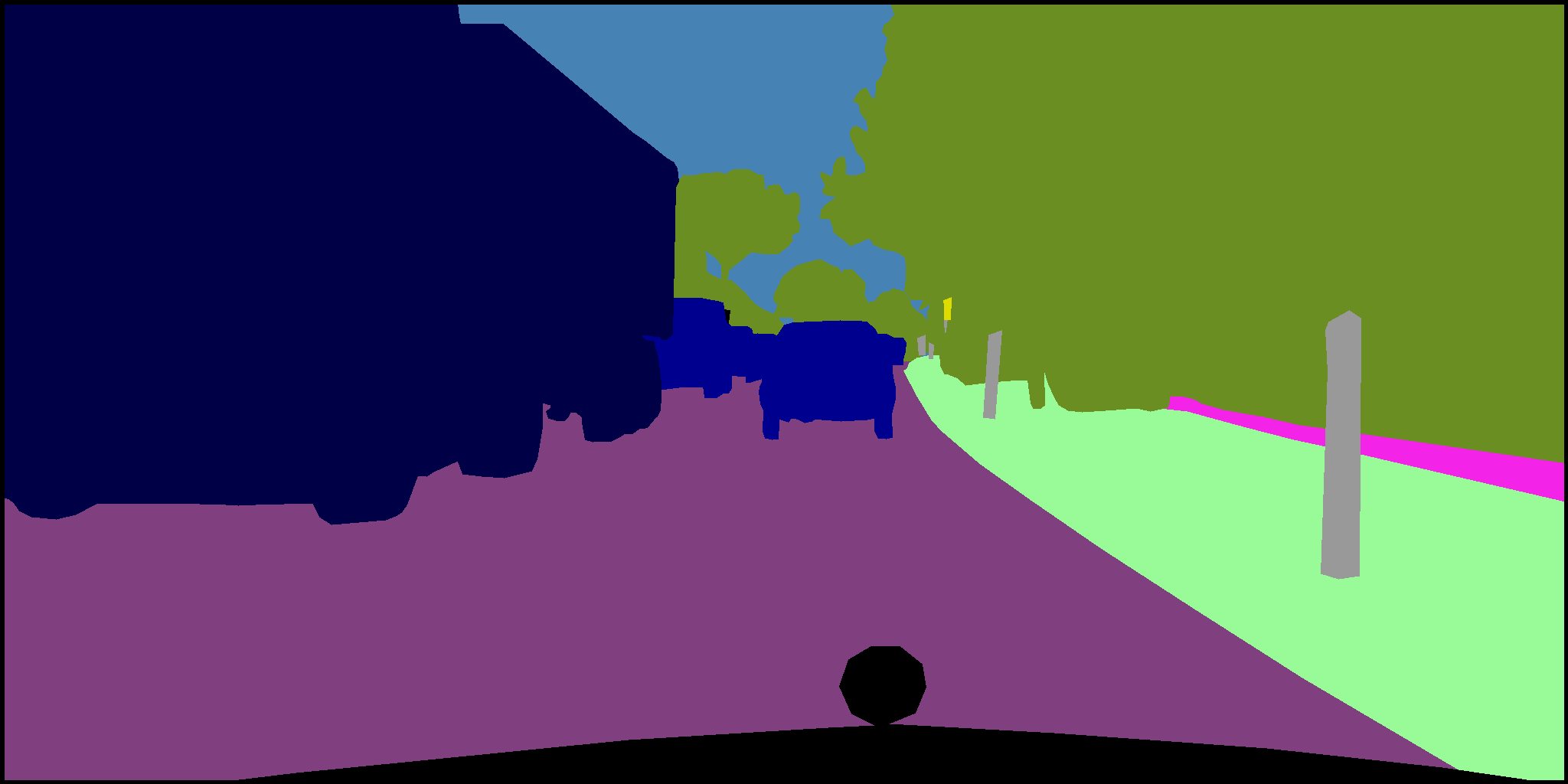}
	\end{subfigure}
	
	\vspace{0.5mm}
   
	\begin{subfigure}{0.16\textwidth}
		\includegraphics[width=\textwidth]{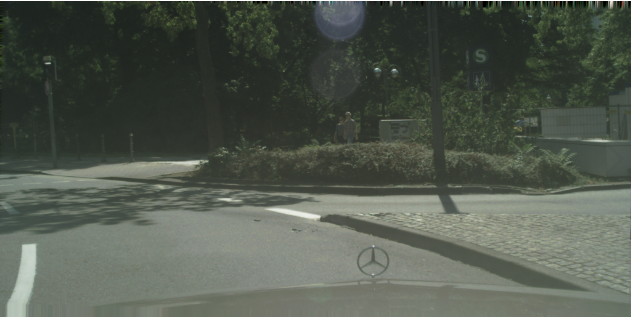}
		\caption{Image}
	\end{subfigure}
	\begin{subfigure}{0.16\textwidth}
		\includegraphics[width=\textwidth]{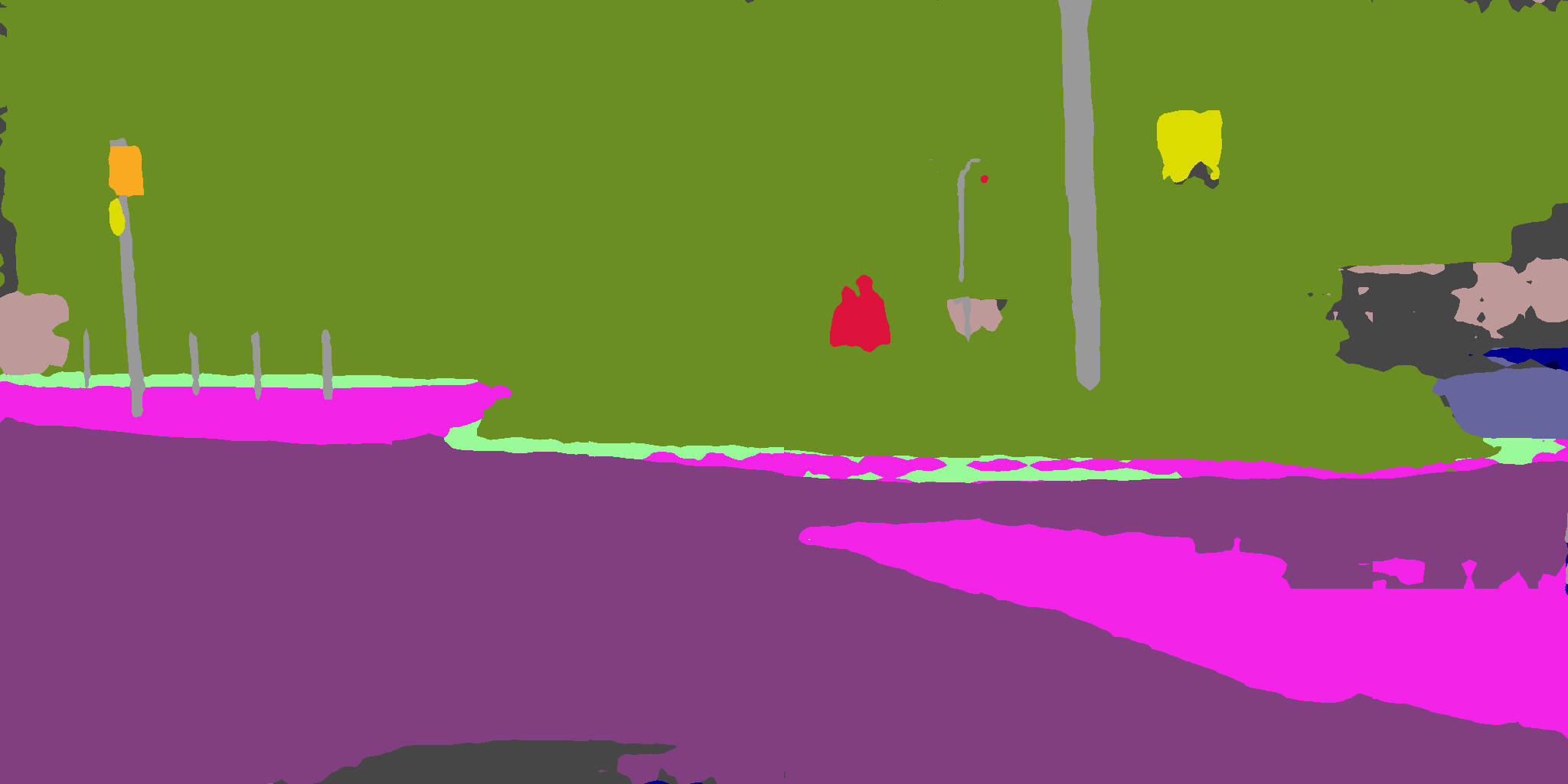}
		\caption{FCN}
	\end{subfigure}
	\begin{subfigure}{0.16\textwidth}
		\includegraphics[width=\textwidth]{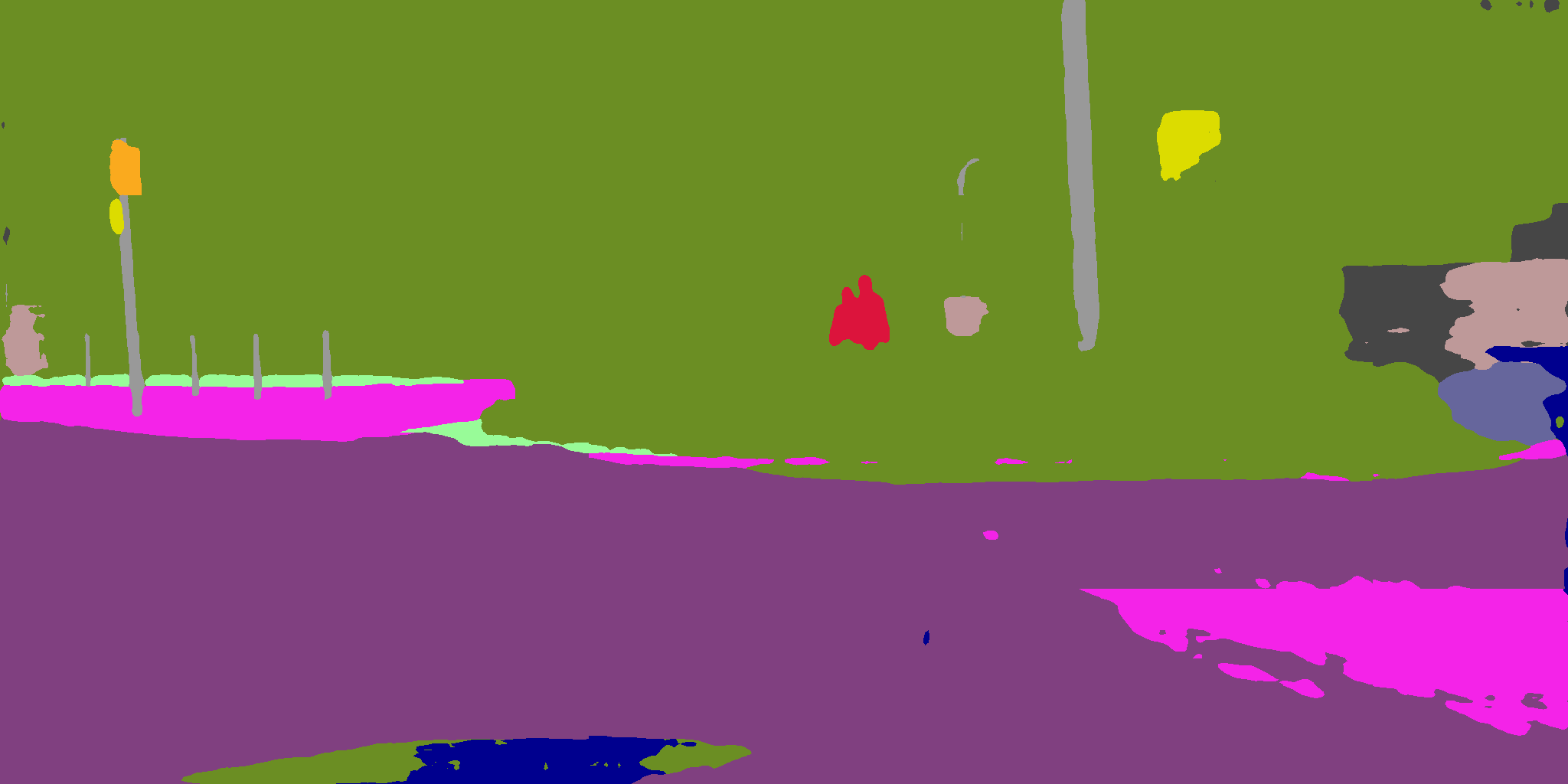}
		\caption{ASPP}
	\end{subfigure}
	\begin{subfigure}{0.16\textwidth}
		\includegraphics[width=\textwidth]{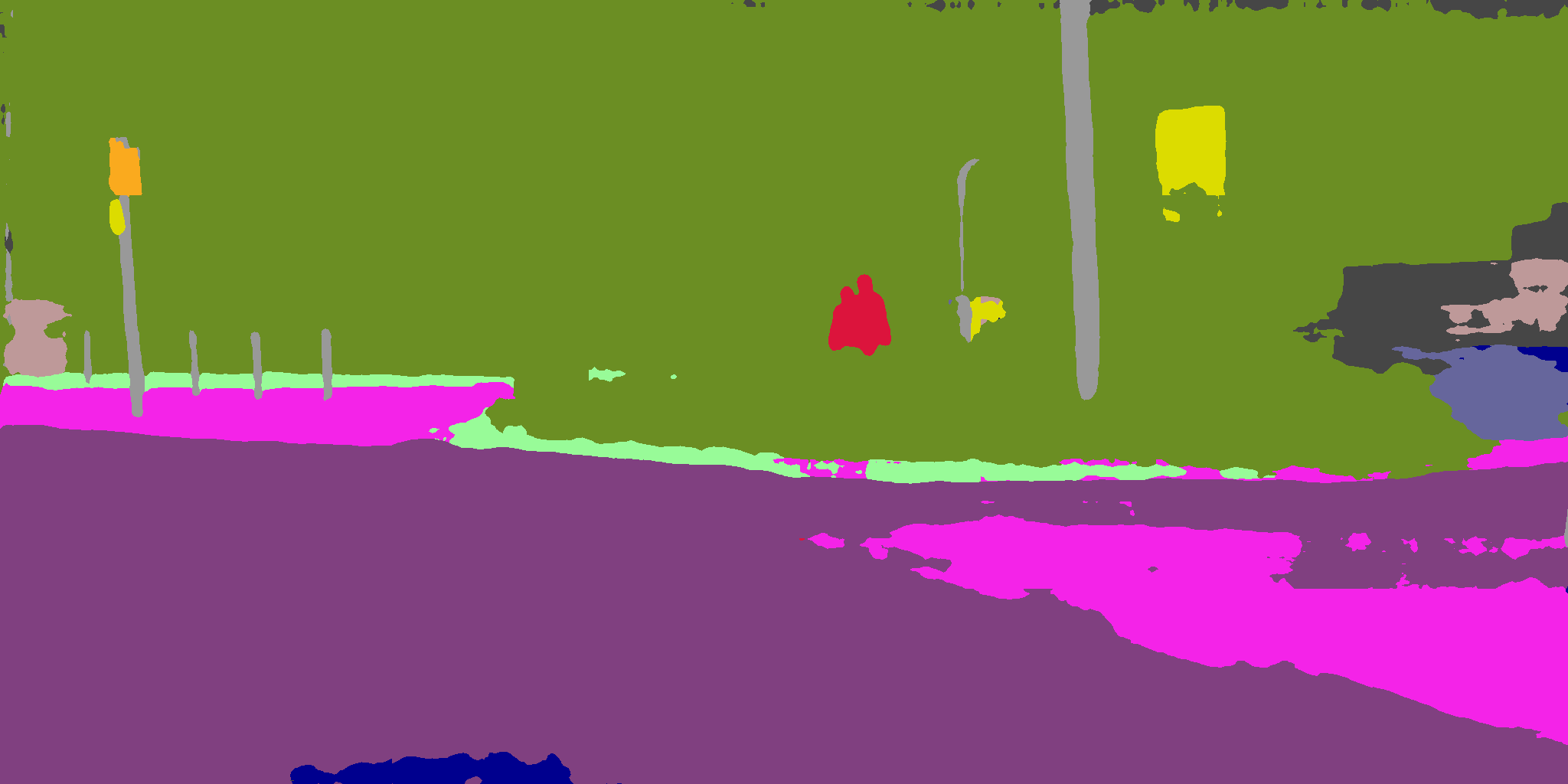}
		\caption{PSP}
	\end{subfigure}
	\begin{subfigure}{0.16\textwidth}
		\includegraphics[width=\textwidth]{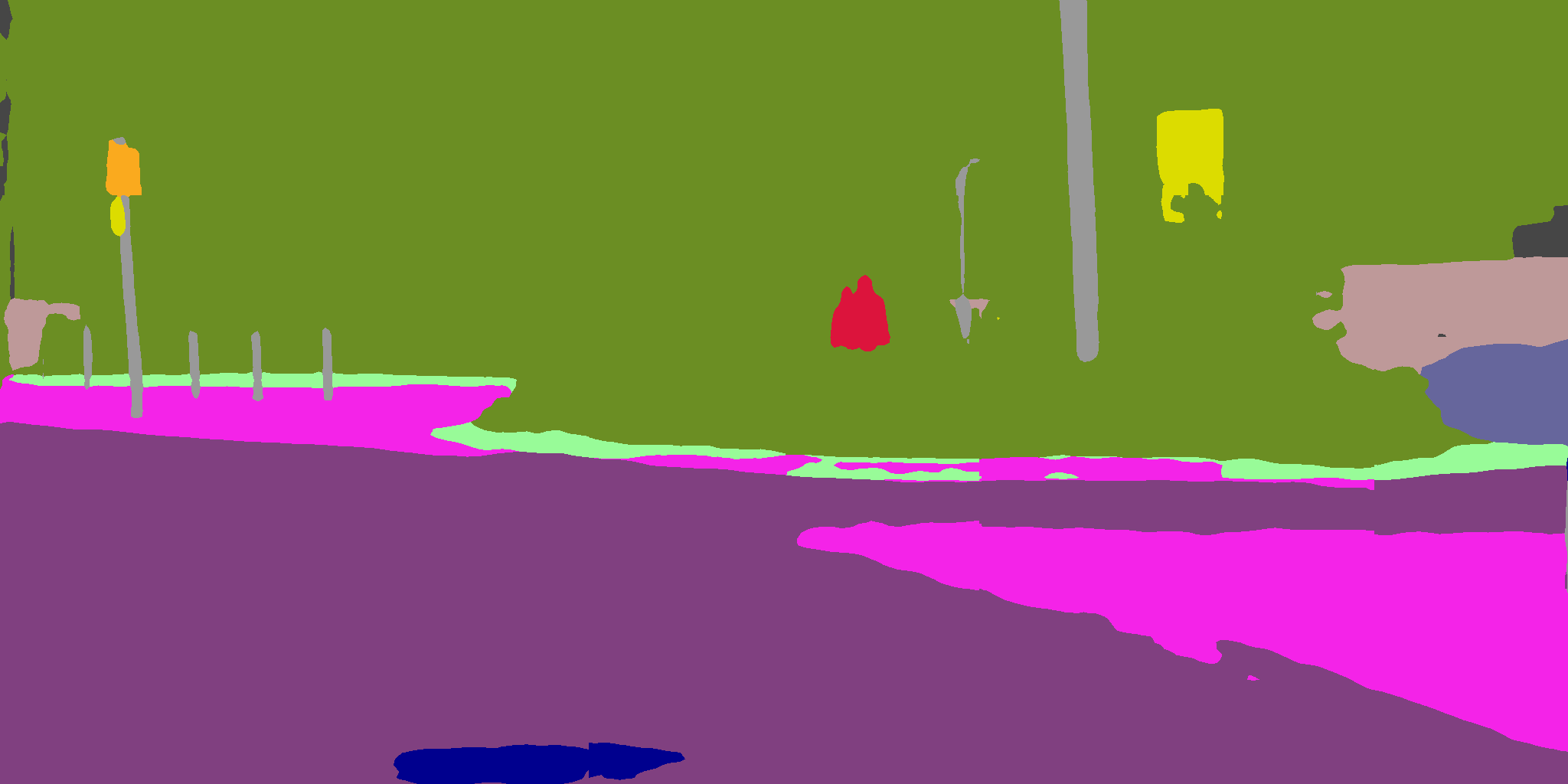}
		\caption{SpyGR}
	\end{subfigure}
	\begin{subfigure}{0.16\textwidth}
		\includegraphics[width=\textwidth]{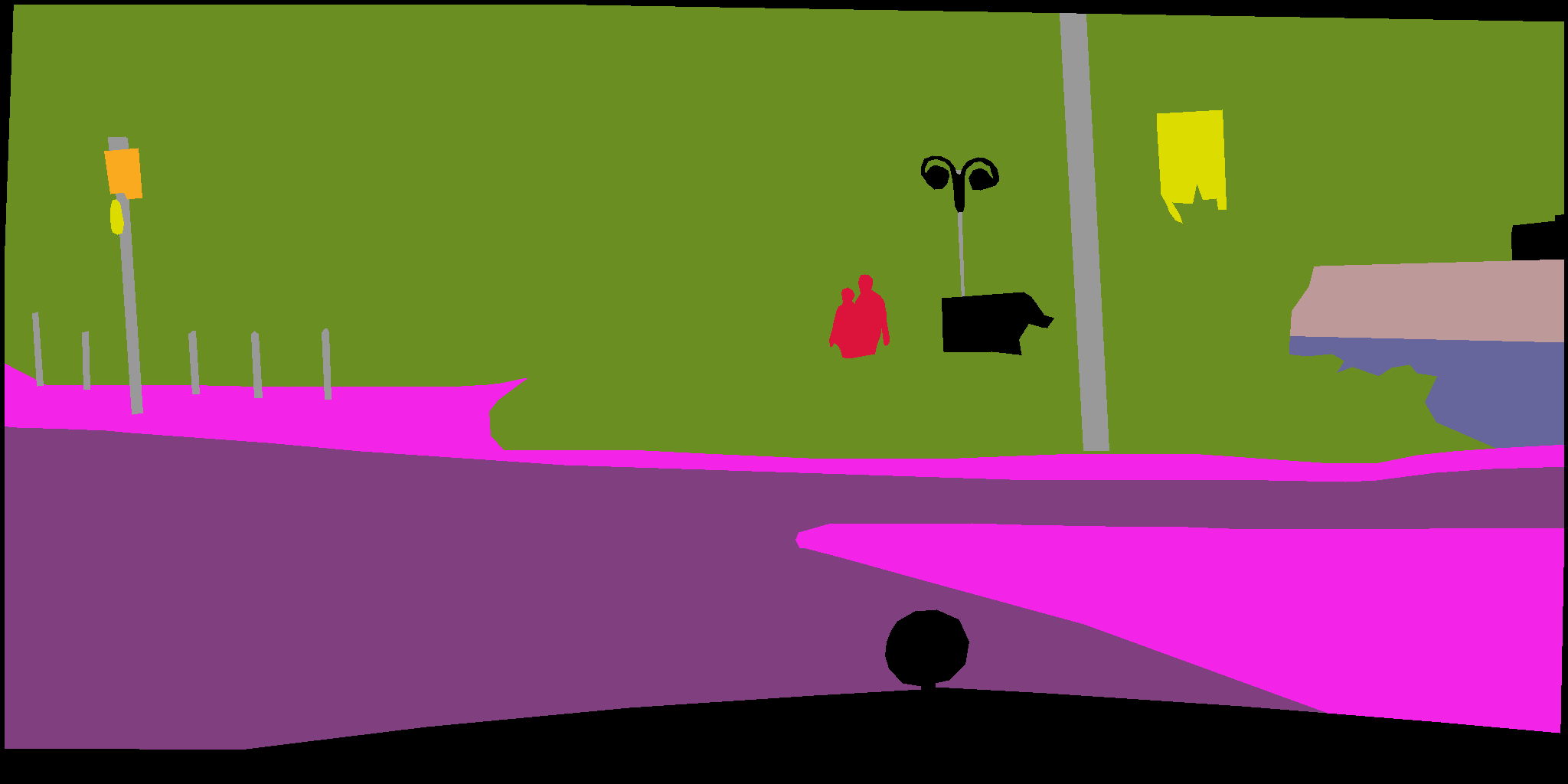}
		\caption{Label}
	\end{subfigure}
	\vspace{-3mm}
	\caption{Visualisation comparison with other methods.}
	\label{fig-vis}
\end{figure*}

\noindent\textbf{With data-independent $\tilde{\Lambda}$.} Corresponding to Eq~(\ref{eq6}), we now introduce a diagonal matrix into the inner product of $\phi$ and $\phi^T$ to have a better distance metric. However, we make the diagonal matrix $\tilde{\Lambda}$ feature independent, which means that it is a vector of parameters to learn. It outperforms the simplest GCN by 0.60. We can see that the diagonal matrix indeed makes a better distance metric with only a few trainable parameters, and leads to a higher performance. 

\noindent\textbf{With data-dependent $\tilde{\Lambda}(X)$.} In this case, we calculate $\tilde{A}$ using Eq~(\ref{eq6}), and the attention diagonal matrix becomes data-dependent by Eq~(\ref{eq7}). This mechanism works in a way similar to soft-attention. As a result, It further has a performance gain of $0.47$ on mIoU over the data-independent case. It is demonstrated that the attention diagonal matrix $\tilde{\Lambda}(X)$ is more representative, and provides a better distance metric conditioned on the distribution of input features. 

\noindent\textbf{Identity.} Now we recover the identity term in the Laplacian formulation, and calculate $\tilde{L}$ exactly following Eq~(\ref{eq5}). The identity term also plays a role of shortcut connection to facilitate optimization of graph reasoning. We see that the performance has a further increment.

\noindent\textbf{Spatial Pyramid.} Finally, we organize the input feature as a spatial pyramid following Eq~(\ref{eq10}), which enables capturing multiple long-range contextual patterns from different scales. It further brings a performance gain of 0.51 in mIoU.

\subsection{Analysis}

In order to have a better sense of the effects of our proposed spatial pyramid based graph reasoning, we visualize the similarity matrix in different scales on the Cityscapes dataset. Concretely, as shown in Figure~\ref{fig-Z}, we randomly generate a sampling point $i$ and mark it by the green cross. And then we visualize the $i$-th row of the similarity matrix, \emph{i.e.}, $\tilde{A}_i\in\mathbb{R}^{H\times W}$, as a heatmap. The right four columns show the similarity matrix from the coarsest level to the finest level. We can observe that, different long-range contextual patterns are captured in the spatial pyramid. For the sampling points located on the car, the strongest activations of the four scales are distributed on different cars. These different long-range relationships are finally aggregated into the finest level for prediction. This also happens to other categories such as sidewalk, bus and vegetation. For the sampling points located on the boundary line of two semantic categories, the interactions in different scales help to better assign the pixel into the right category. The aforementioned analysis shows that our proposed spatial pyramid is able to aggregate rich semantic information and capture multiple long-range contextual patterns. We also show the visualisation comparison with other methods in Figure~\ref{fig-vis}.

\section{Conclusion}

In this paper, we aim to model long-range context using graph convolution for the semantic segmentation task. Different from current methods, we perform our graph reasoning directly in the original feature space organized as a spatial pyramid. We propose an improved Laplacian that is data-dependent, and introduce an attention diagonal matrix on the inner product to make a better distance metric. Our method gets rid of projecting and re-projecting processes, and retains the spatial relationships that enables spatial pyramid. We adopt a computing scheme to reduce the computational overhead significantly. Our experiments show that each part of our design contributes to the performance gain, and we outperform other methods without introducing more computational or memory consumption. 


\section{Acknowledgement}
Zhouchen Lin is supported by National Natural Science Foundation
(NSF) of China~(grant no.s 61625301 and 61731018), Major Scientific Research Project of Zhejiang Lab (grant no.s 2019KB0AC01 and 2019KB0AB02), Beijing Academy of Artificial Intelligence, and Qualcomm. Hong Liu is supported by NSF China~(grant no. U1613209) and NSF Shenzhen~(grant no. JCYJ20190808182209321).

{\small
\bibliographystyle{ieee_fullname}
\bibliography{egbib}
}

\end{document}